\documentclass[sigconf]{acmart}

\AtBeginDocument{%
  }

\setcopyright{acmlicensed}
\copyrightyear{2018}
\acmYear{2018}
\acmDOI{XXXXXXX.XXXXXXX}
\acmConference[Conference acronym 'XX]{Make sure to enter the correct
  conference title from your rights confirmation email}{June 03--05,
  2018}{Woodstock, NY}
\acmISBN{978-1-4503-XXXX-X/2018/06}

\usepackage{amsmath,mathtools}
\usepackage{algorithm}
\usepackage{algpseudocode}
\usepackage{booktabs}
\usepackage{tcolorbox}
\tcbset{promptbox/.style={colback=gray!5!white, colframe=blue!75!black, boxrule=0.5pt, arc=2mm, left=1mm, right=1mm, top=1mm, bottom=1mm}}
\usepackage{multirow}
\usepackage{graphicx}
\usepackage{colortbl}
\usepackage{adjustbox}
\usepackage{makecell}
\usepackage{subcaption}
\usepackage{booktabs}
\usepackage{tabularx}
\usepackage{array}
\usepackage{makecell}
\usepackage{xcolor}
\newcommand{\eststd}[2]{%
    #1\hspace{0.05em}%
    \smash{%
        \raisebox{-0.55ex}{%
            \textcolor{black!55}{\scriptsize(#2)}%
        }%
    }%
}

\begin{document}

\title{Different Feedback, Different Updates: Selective Self-Learning from User Interactions for Large Language Models}
\author{Xuancheng Li}
\authornote{This work was done during an internship at Tencent.}
\email{lixuancheng23@mails.tsinghua.edu.cn}
\affiliation{%
  \institution{DCST, Tsinghua University}
  \city{Beijing}
  \country{China}
}
\author{Haitao Li}
\email{liht22@mails.tsinghua.edu.cn}
\affiliation{%
  \institution{DCST, Tsinghua University}
   \city{Beijing}
  \country{China}
}

\author{Yujia Zhou}
\email{zhouyujia@mail.tsinghua.edu.cn}
\affiliation{%
  \institution{DCST, Tsinghua University}
     \city{Beijing}
  \country{China}
}

\author{Qingyi Pan}
\email{panqingyi19@gmail.com}
\affiliation{%
  \institution{DCST, Tsinghua University}
   \city{Beijing}
  \country{China}
}

\author{Heng Wang}
\email{ryzeewang@tencent.com}
\affiliation{%
  \institution{Tencent}
   \city{Beijing}
  \country{China}
}

\author{Yiqun Liu}
\email{yiqunliu@tsinghua.edu.cn}
\affiliation{%
  \institution{DCST, Tsinghua University}
     \city{Beijing}
  \country{China}
}

\author{Min Zhang}
\email{z-m@tsinghua.edu.cn}
\affiliation{%
  \institution{DCST, Tsinghua University}
     \city{Beijing}
  \country{China}
}

\author{Qingyao Ai}
\email{aiqingyao@gmail.com}
\affiliation{%
  \institution{DCST, Tsinghua University}
     \city{Beijing}
  \country{China}
}

\providecommand{\method}{\textsc{SLIFT}}
\providecommand{\methodfull}
{Self-Learning from Interaction Feedback via Task-Relative Specialization}
\providecommand{\methodfullbf}
{\textbf{S}elf-\textbf{L}earning from \textbf{I}nteraction \textbf{F}eedback via \textbf{T}ask-Relative Specialization}

\begin{abstract}
User feedback offers natural supervision for persistent LLM improvement, but
a single message may support multiple behavioral changes with different
scopes of generalization. We
introduce \textbf{\method{}}, a selective self-learning framework built on a
\emph{task-relative view} of user feedback. \method{} decomposes each feedback
message into atomic components and interprets each component relative to the
original task as \textsc{Fix}, \textsc{Spec}, or \textsc{Null}: requirements
for task validity, compatible condition-specific refinements, or content with
no reliable positive update direction. 
To incorporate each change at the appropriate scope, \method{} trains two
complementary LoRA adapters on a shared frozen backbone: a Generalist that
consolidates \textsc{Fix} requirements into default behavior through
feedback-conditioned self-distillation, and a Specialist that observes only
the task and Generalist response to supply residual guidance for applicable,
unmet \textsc{Spec} refinements. \textsc{Null} components induce no positive
update. Across backbones, \method{} achieves strong performance on both MemoryBench and WildFB, with targeted analyses further examining its underlying mechanisms. \footnote{We release our code at
\url{https://anonymous.4open.science/r/SLIFT}.}
\end{abstract}



\maketitle

\section{Introduction}
\label{sec:intro}
Large language models (LLMs) are increasingly deployed in interactive systems, where users respond to model outputs across turns.
These responses often reveal errors, unmet requirements, or other shortcomings in the preceding outputs, implicitly signaling how the model should have responded instead. Recent studies have shown that feedback embedded in interaction histories provides useful supervision for model improvement \cite{chen-etal-2025-retrospective,pmlr-v235-tucker24a,shi-etal-2026-wildfeedback}. As user--LLM interactions continue to scale, interaction logs are becoming an increasingly important source of such supervision.

Models can often exploit this supervision within the interaction in which it
appears. When user feedback provides an actionable signal, a model can identify problems in its preceding response and produce a more appropriate revision \cite{akyurek-etal-2023-rl4f}. This suggests that models can interpret the feedback
in context and adjust their behavior accordingly. However, without a persistent update, this adaptation remains confined to the current dialogue and is not retained once the feedback is no longer available \cite{yan-etal-2024-refutebench}.
This motivates the problem of transforming interaction-specific hindsight into persistent behavioral improvements that appropriately generalize to future interactions \cite{liu2023chainhindsightalignslanguage}.

Realizing this objective first requires determining what behavioral change, if any, is supported by each feedback message. This is challenging because natural user feedback rarely takes the form of explicit structured supervision \cite{chen-etal-2025-retrospective}. Real-world user feedback is highly diverse \cite{zheng2024lmsyschat1mlargescalerealworldllm,zhao2024wildchat1mchatgptinteraction}: a single feedback may itself combine corrections, preferences, and new requests \cite{liu-etal-2025-user}. Because these components may warrant different behavioral changes, treating the entire feedback as a single supervision signal can obscure what should be learned and how broadly it should generalize. Learning from user feedback therefore requires distinguishing these components.

Existing approaches to learning from user feedback broadly pursue persistent
adaptation through two routes: non-parametric memory and parametric updates.
Non-parametric methods store interaction-derived experience outside model
parameters and retrieve it for context-dependent reuse
\cite{kang-etal-2025-memory,xu2025amemagenticmemoryllm}, whereas parametric methods encode
interaction-derived behavioral changes directly into model parameters.
Within parametric methods, a central challenge is converting raw feedback
into a trainable signal, since feedback often specifies neither a complete
target response nor an explicit scalar reward. Some approaches convert feedback into scalar rewards or preference-based objectives
\cite{ouyang2022training,peng-etal-2026-wildreward,
li2026mulferlenhancingreinforcementlearning}, while a separate approach
distills a hindsight policy conditioned on the full feedback message
\cite{kleinebuening2026aligning}. These approaches preserve different amounts
of feedback structure. Scalar rewards and preference objectives compress rich
verbal feedback into coarse outcome or comparative signals, which may obscure
fine-grained guidance
\cite{li2026mulferlenhancingreinforcementlearning}. By contrast, hindsight
distillation retains the full feedback message but may provide weak
supervision when the model cannot reliably interpret and act on it
\cite{kleinebuening2026aligning}.

Another challenge is deciding how broadly each feedback-driven change should generalize. Some feedback identifies a property needed to satisfy the
original task and may therefore support a reusable change to default behavior.
Other feedback expresses a compatible refinement whose relevance depends on
the task or the response already produced. Generalizing such refinements as
default behavior could propagate a context-specific change to unsupported
contexts \cite{stephan2024rlvf}. A subsequent user feedback may also contain
unrelated requests, conflicting instructions, or ambiguous content that
provides no reliable positive target. Because a signal's learning role depends
on its relation to the original task rather than its surface form, the model
must decide whether the supported change should be consolidated, remain
conditional, or drive no positive update.

Together, these challenges require determining what should be learned from
each feedback component and how broadly the resulting change should
generalize. We therefore ask:
\emph{How can a model achieve continual self-improvement from user
interactions while incorporating each feedback-supported change within its
appropriate scope?}

\begin{figure}[t]
\centering
\includegraphics[width=\columnwidth]{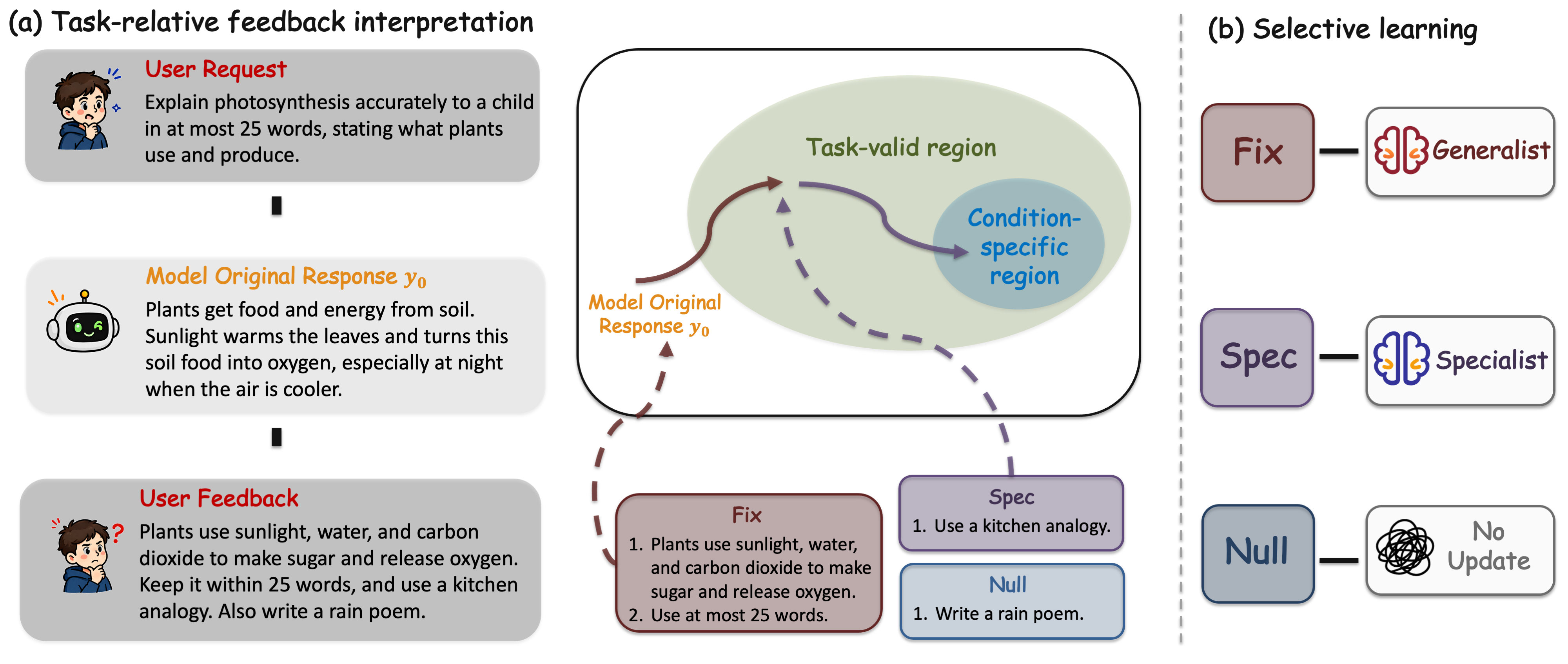}
\caption{Illustration of task-relative feedback decomposition. A composite
feedback message is decomposed into atomic \textsc{Fix}, \textsc{Spec},
and \textsc{Null} components relative to the original task.}
\label{fig:overview}
\vspace{-2.0em}
\end{figure}

To address this question, we introduce \textbf{\method{}}
(\methodfullbf{}), a selective self-learning framework built on a
\emph{task-relative view} of user feedback. Under this view, the
original task defines the task-valid region and serves as the semantic
reference for interpreting subsequent feedback. As illustrated in
Figure~\ref{fig:overview}, the three feedback roles are defined by how
each component relates to this region. Operationally, \method{} uses
its frozen backbone to decompose each subsequent user turn into atomic
components and assigns each component a role using a fixed three-way
rubric. A \textsc{Fix} specifies a property that every response in the
task-valid region must satisfy; a \textsc{Spec} specifies a compatible
refinement that selects a condition-specific subregion within it; and
a \textsc{Null} provides no reliable positive update direction for
the original task. Together, these roles distinguish requirements of
task validity, condition-specific refinements among task-valid
responses, and content that should not drive a positive update.

Because these roles support changes with different scopes, \method{}
maps them to complementary learning pathways rather than a single
undifferentiated update. Applying \textsc{Spec} supervision through
the same pathway as \textsc{Fix} could turn conditional evidence into
default behavior and extend it beyond relevant contexts
\cite{stephan2024rlvf}. Inspired by functional specialization in
complementary learning systems
\cite{mcclelland1995complementary}, we instantiate two LoRA \cite{hu2021loralowrankadaptationlarge} adapters
on a shared frozen backbone. The Generalist learns from \textsc{Fix}
components through feedback-conditioned self-distillation,
consolidating task-necessary changes into its default behavior. The
Specialist observes only the task and the Generalist response; using
targets constructed from \textsc{Spec} components, it learns
whether a compatible refinement applies and remains unmet and what residual guidance is needed. \textsc{Null} components provide no
positive update. Together, these pathways separate default-behavior
consolidation from the selective refinement of otherwise task-valid
responses. Supervision for both pathways is constructed by the frozen
backbone using the corresponding feedback components, without
requiring a stronger teacher or explicit reward model.

Our contributions are threefold.
First, we introduce a task-relative formulation that decomposes feedback
into atomic \textsc{Fix}, \textsc{Spec}, and \textsc{Null} components.
Second, we propose \method{}, whose frozen backbone constructs supervision
for two pathways: the Generalist consolidates \textsc{Fix} requirements
into default behavior, while the Specialist supplies residual guidance only
for applicable, unmet \textsc{Spec} refinements.
Third, across backbones, \method{} achieves strong performance on both MemoryBench~\cite{ai2026memorybenchbenchmarkmemorycontinual} and WildFB~\cite{peng-etal-2026-wildreward}, while targeted analyses further probe its mechanisms.

\section{Related Work}

\subsection{User Feedback as a Learning Resource}

Large-scale corpora such as LMSYS-Chat-1M
~\cite{zheng2024lmsyschat1mlargescalerealworldllm} and WildChat
~\cite{zhao2024wildchat1mchatgptinteraction} provide naturally occurring user--LLM
interaction logs, where feedback is often implicit, composite, and
noisy~\cite{liu-etal-2025-user}. Recent resources convert such logs
into training signals: WildFeedback
~\cite{shi-etal-2026-wildfeedback} constructs preference pairs,
WildFB~\cite{peng-etal-2026-wildreward} provides ordinal satisfaction
labels, and MemoryBench~\cite{ai2026memorybenchbenchmarkmemorycontinual} offers simulated
explicit and implicit feedback for training and held-out evaluation.
These resources generally preserve feedback at the turn level or
reduce it to an aggregate signal. We instead decompose each feedback
turn into atomic, task-relative components that distinguish
task-necessary changes, conditional refinements, and content that
should not drive a positive update.

\subsection{Continual Learning from User Feedback}
\label{sec:2_2}

Continual learning from interaction feedback follows two routes:
non-parametric memory and parametric updates. MemoryOS
~\cite{kang-etal-2025-memory}, A-Mem~\cite{xu2025amemagenticmemoryllm},
MemOS~\cite{li2025memos}, and ReMem~\cite{wei2025evomemory} store
interaction-derived experience externally for later retrieval or
refinement. Parametric methods instead turn logs into persistent
updates: ReSpect~\cite{chen-etal-2025-retrospective} learns
retrospectively, IEvoAgent~\cite{cai-etal-2026-ievoagent} estimates
feedback-based rewards,
WildReward~\cite{peng-etal-2026-wildreward}
learns ordinal satisfaction signals, and SDPO~\cite{kleinebuening2026aligning} distills
a feedback-conditioned hindsight distribution. In verifiable domains, SEAM~\cite{li-etal-2026-beyond-experience}
optimizes structured guidance through executor rollouts, while
MulFeRL~\cite{li2026mulferlenhancingreinforcementlearning} learns from
feedback-guided regeneration. FCP
~\cite{luo2025languagemodelslearnverbal} conditions generation on desired verbal
feedback, which remains an input at test time, while C3PO
~\cite{stephan2024rlvf} learns where a predefined high-level feedback
rule should apply. In contrast, \method{} decomposes naturally
occurring feedback into atomic, task-relative components and maps them
to consolidation into default behavior, conditional refinement, or no
positive update.

\section{Method}
\label{sec:method}
\begin{figure*}[!t]
    \centering
    \includegraphics[width=\textwidth]{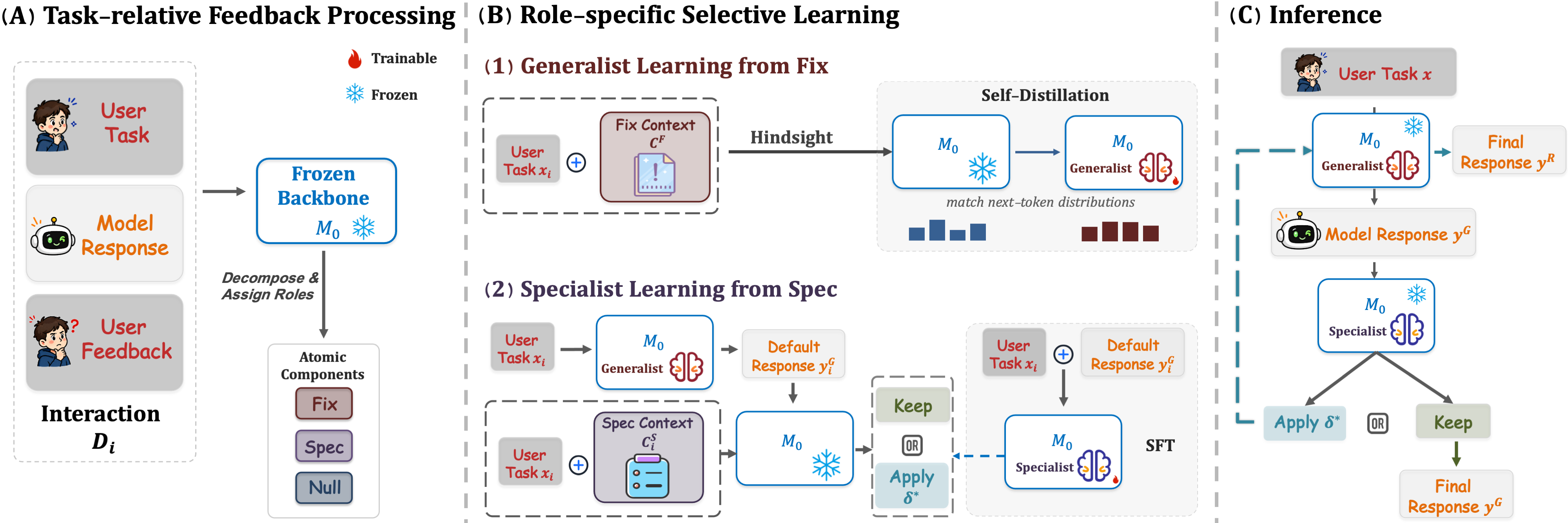}
    \caption{Illustration of SLIFT.}
    \label{fig:main}
\end{figure*}

\subsection{Problem Formulation}
\label{sec:problem}

Consider a language model $M_0$ and a collection of $N$ logged
interactions:
\begin{equation}
\mathcal D
=
\left\{
d_i
\right\}_{i=1}^{N},
\qquad
d_i
=
\left(
x_i,
y_i^{\mathrm{log}},
u_i
\right),
\label{eq:interaction}
\end{equation}
where $x_i$ denotes the user task, including any preceding context
needed to specify the user's request; $y_i^{\mathrm{log}}$ is the
logged assistant response; and $u_i$ is the subsequent user turn, which
may contain feedback on that response.

When $u_i$ is available in context, a model may infer how to improve its
preceding response. However, this improvement is not retained once the
feedback is removed. Moreover, a single user turn may imply multiple
behavioral changes that apply under different task conditions, or may provide
no reliable basis for persistent adaptation. The problem is therefore not
only what should be learned from an interaction, but also under which
conditions each learned change should apply.
This leads to our central research question:
\emph{How can a model achieve continual self-improvement from user
interactions while incorporating each feedback-supported change within its
appropriate scope?}

Given $\mathcal D$, we seek an updated model $M_\Theta$ that retains
feedback-supported behavioral changes on future interactions within
their appropriate scope, while avoiding unsupported changes outside
that scope. We next realize this objective by recovering task-relative learning signals and incorporating them through selective updates.

\subsection{Method Overview}
\label{sec:overview}

To realize the objective above, we introduce \method{}
(\emph{\methodfull}). As illustrated in Figure~\ref{fig:main},
\method{} proceeds in two stages: task-relative feedback processing
and selective learning. In the first stage, following the
task-relative view introduced in \S~\ref{sec:intro}, the frozen
backbone decomposes each subsequent user feedback into atomic components
and assigns each a role relative to the original task:
\textsc{Fix} for task-necessary properties, \textsc{Spec} for
compatible but non-necessary refinements, and \textsc{Null} when no
reliable positive behavioral target can be recovered.

In the second stage, these roles determine the form of adaptation.
A \textsc{Fix} specifies a requirement shared by all task-valid responses,
allowing the Generalist to consolidate the corresponding change into its
default behavior.
A \textsc{Spec}, by contrast, selects among otherwise task-valid
responses: whether it applies depends on the observable task context,
while whether it still requires action depends on the Generalist
response. Consolidating \textsc{Spec} supervision through the same
pathway would turn conditional evidence into default behavior and risk
applying a locally supported refinement outside its appropriate context
\cite{stephan2024rlvf}. The Specialist therefore learns from the
task--response pair whether a refinement is needed and, if so, what
remains to be changed. \textsc{Null} components provide no positive
learning target.

Inspired by the functional specialization of complementary learning
systems
\cite{mcclelland1995complementary}, we instantiate separate
Generalist and Specialist LoRA adapters on a shared frozen backbone.
This separation provides an inductive bias over update scope. The
Generalist learns \textsc{Fix}-derived task requirements through
feedback-conditioned self-distillation, consolidating them into its
default behavior. The Specialist is trained via supervised fine-tuning on
$(x_i,y_i^G)$, where $y_i^G$ is the frozen Generalist's response and
the corresponding \textsc{Spec} components are used to construct
the target: \texttt{KEEP} if no applicable refinement remains unmet,
or \texttt{APPLY} with residual guidance otherwise.

At inference time, the Generalist produces a default response. The Specialist either preserves it or generates residual guidance; only in the latter case does the frozen Generalist perform a single residual-guided integration pass.

\subsection{Task-Relative Feedback Processing}
\label{sec:projection}

The first stage separates what a feedback turn expresses from how its
content should update the model. Using the frozen backbone $M_0$,
\method{} first decomposes the user feedback into atomic
components and then interprets each component relative to the original
task as supporting a reusable update to default behavior, a
conditional refinement, or no positive update.

\textit{Atomic feedback extraction.}
A user feedback $u_i$ may combine distinct signals, such as a
correction and a separate request
\cite{chen-etal-2025-retrospective,
shi-etal-2026-wildfeedback,
liu-etal-2025-user}.
We use $M_0$ to decompose it into minimal, self-contained components:
\begin{equation}
\mathcal C_i
=
\operatorname{Extract}_{M_0}
\left(
x_i,
y_i^{\mathrm{log}},
u_i
\right)
=
\left(
c_{i1},
\ldots,
c_{im_i}
\right).
\label{eq:feedback-extraction}
\end{equation}
Each component captures one candidate learning signal; its role is
determined only in the next step. If no self-contained signal can be
isolated, then $\mathcal C_i=\varnothing$.

\textit{Task-relative interpretation.}
The original task anchors this interpretation by defining the space of
fully valid responses. As illustrated in
Figure~\ref{fig:overview}, let $\mathcal V(x_i)$ denote the responses
that fully satisfy the requirements inferable from $x_i$. We first check whether a component names a specific, compatible property of the requested deliverable. Components that do not are
  labeled \textsc{Null}—for example, those that are ambiguous, irrelevant, conflicting, or that raise a separate request. For each remaining component, define:
\begin{equation}
\mathcal V(x_i;c_{ij})
=
\left\{
y\in\mathcal V(x_i)
\mid
y\models c_{ij}
\right\},
\label{eq:component-region}
\end{equation}
where $y\models c_{ij}$ means that $y$ satisfies the property expressed
by $c_{ij}$.

The omission criterion then distinguishes the two positive learning
roles. If
$\mathcal V(x_i;c_{ij})=\mathcal V(x_i)$, every fully valid response
must satisfy the component; omitting it precludes full task validity,
so it is assigned \textsc{Fix}. If
$\varnothing\neq\mathcal V(x_i;c_{ij})\subsetneq\mathcal V(x_i)$,
the component selects a compatible subregion of otherwise task-valid
responses; omitting it may preserve full validity, so it is assigned
\textsc{Spec}. Thus, a \textsc{Fix} identifies a condition of task
validity, whereas a \textsc{Spec} refines the choice among task-valid
responses. In Figure~\ref{fig:overview}, the factual correction and
length constraint are \textsc{Fix}, the kitchen analogy is
\textsc{Spec}, and the unrelated poem request is \textsc{Null}.
Treating a \textsc{Spec} as default behavior could extend a locally
supported refinement beyond its appropriate context
\cite{stephan2024rlvf}.

Prior work similarly shows that implicit feedback signals
can be recognized from interactions and that in-situ user feedback can
be identified and classified from conversation logs
\cite{chen-etal-2025-retrospective,
shi-etal-2026-wildfeedback}.
Operationally, we present $(x_i,c_{ij})$ to $M_0$ with a fixed
three-way rubric implementing these criteria. Because extraction has
already isolated a single candidate signal, role assignment becomes a
constrained classification task rather than open-ended feedback
discovery. \textsc{Fix} components supervise the Generalist's default
behavior, \textsc{Spec} components are used to construct
response-conditioned Specialist targets, and \textsc{Null} components
provide no positive behavioral target.

\subsection{Generalist Learning from \textsc{Fix} Components}
\label{sec:generalist}

As discussed in \S\ref{sec:overview}, the Generalist consolidates
\textsc{Fix}-derived task requirements into its default behavior,
whereas the Specialist handles optional refinements through selective
intervention.

For each interaction $d_i$, we concatenate all components $c_{ij}\in\mathcal C_i$ assigned \textsc{Fix} relative to the task $x_i$ in their original order. We denote the resulting sequence by $C_i^F$. Interactions with $C_i^F\neq\varnothing$ form the Generalist training set $\mathcal D_F$.

Because $C_i^F$ specifies task-validity requirements rather than a
complete target response, synthesizing a full revision would require an
additional generation pass and turn incidental rewriting choices into
token-level supervision. Model-based rewriting is also prone to
over-editing content that should remain unchanged
\cite{adams-etal-2022-learning,zeng-etal-2026-hyperedit}. For example,
correcting a factual error may unnecessarily alter organization, detail,
or style, thereby biasing learning toward properties unsupported by the
observed feedback. We therefore
adapt Self-Distillation Policy Optimization (SDPO) \cite{hubotter2026reinforcement,kleinebuening2026aligning}, a
feedback-conditioned self-distillation method.
This converts the identified requirements into dense token-level
supervision while avoiding revised-response construction and the need
for an external teacher or explicit reward model.

SDPO was originally formulated as an on-policy algorithm: the current
policy samples a response, receives feedback, and then re-evaluates the
same response through a feedback-conditioned self-teacher
\cite{hubotter2026reinforcement}. However, user interaction data are often available as pre-collected logs, where responses already exist and need not have been generated by the policy being optimized \cite{kleinebuening2026aligning}. Building on the offline extension of SDPO to logged interactions, we tailor it to selective \textsc{Fix} learning by treating $y_i^{\mathrm{log}}$ as a fixed rollout and using only $C_i^F$ as hindsight context. At each prefix
$y_{i,<t}^{\mathrm{log}}$, the frozen backbone produces the hindsight
distribution $M_0(\cdot\mid x_i,C_i^F,y_{i,<t}^{\mathrm{log}})$, which the
Generalist learns to reproduce without observing $C_i^F$.\footnote{We
provide a theoretical justification for this offline approximation in
\S~\ref{app:offline-sdpo}.}

Replacing on-policy rollouts with logged responses yields an
off-policy variant of SDPO. To avoid altering behavior unrelated to the
identified \textsc{Fix} requirements, we add a feedback-free
behavioral anchor. Let $M_G=M_0\oplus A_G$, where $M_0$ is the frozen backbone and
$A_G$ is a trainable LoRA adapter. At each logged prefix, let
$p_{i,t}^{G}=M_G(\cdot\mid x_i,y_{i,<t}^{\mathrm{log}})$,
$p_{i,t}^{H}=M_0(\cdot\mid x_i,C_i^F,y_{i,<t}^{\mathrm{log}})$, and
$p_{i,t}^{B}=M_0(\cdot\mid x_i,y_{i,<t}^{\mathrm{log}})$
denote the Generalist, hindsight, and feedback-free distributions,
respectively. We minimize:
\begin{equation}
\mathcal L_G(A_G)
=
\mathbb E_{d_i\sim\mathcal D_F}
\frac{1}{|y_i^{\mathrm{log}}|}
\sum_{t=1}^{|y_i^{\mathrm{log}}|}
\left[
D_{\mathrm{KL}}
\left(
p_{i,t}^{G}
\middle\|
p_{i,t}^{H}
\right)
+
\lambda_B
D_{\mathrm{KL}}
\left(
p_{i,t}^{G}
\middle\|
p_{i,t}^{B}
\right)
\right].
\label{eq:generalist-loss}
\end{equation}

The first term distills the \textsc{Fix}-conditioned hindsight
distribution, while the $\lambda_B$-weighted second term anchors the
Generalist to its feedback-free behavior. We normalize the loss by
response length and update only $A_G$; the resulting $M_G$ is then
frozen to generate the default response $y_i^G$.

\subsection{Specialist Learning from \textsc{Spec} Components}
\label{sec:specialist}

The two pathways separate changes that should become default behavior
from refinements that should remain conditional. As described in
\S~\ref{sec:generalist}, the Generalist learns to reproduce
feedback-conditioned behavior without observing the feedback, thereby
incorporating the induced change into its default behavior. This is
appropriate for task-necessary \textsc{Fix} requirements. A
\textsc{Spec}, however, supports a refinement for the observed
task--response pair but does not directly specify when it should
transfer to future tasks or whether a future response already satisfies
it. Training \textsc{Spec} components through the same pathway would
therefore risk extending a locally supported change beyond its
appropriate scope \cite{stephan2024rlvf}.

To make these decisions, the Specialist acts after the Generalist
produces its default response $y_i^G$, using only the observable
task--response pair $(x_i,y_i^G)$. From this pair, it predicts whether
any supported refinement both applies to the current task and remains
unmet in the response. It outputs \texttt{KEEP} when no such refinement
requires action; otherwise, it outputs \texttt{APPLY} with residual
guidance only for the unmet refinements.

For interaction $i$, let $C_i^S$ denote the possibly empty ordered
sequence of \textsc{Spec} components assigned in
\S~\ref{sec:projection}. During offline target construction, the frozen
backbone receives $(x_i,y_i^G,C_i^S)$, with $C_i^S$ used only as
privileged information. It constructs a target action
$a_i^\star\in\{\texttt{KEEP},\texttt{APPLY}\}$ and, for
\texttt{APPLY}, residual guidance $\delta_i^\star$ describing only the
compatible changes that are still missing from $y_i^G$. The resulting
targets supervise the Specialist from $(x_i,y_i^G)$ alone.

As in \S~\ref{sec:generalist}, we implement the Specialist by attaching
a separate trainable LoRA adapter $A_S$ to the frozen backbone:
$M_S=M_0\oplus A_S$. Each structured target
$z_i^\star=(a_i^\star,\delta_i^\star)$ is serialized as:
\begin{equation}
\operatorname{ser}(z_i^\star)=
\begin{cases}
\texttt{KEEP},
& a_i^\star=\texttt{KEEP},\\
\texttt{APPLY}\,\Vert\,\delta_i^\star,
& a_i^\star=\texttt{APPLY},
\end{cases}
\label{eq:specialist-serialization}
\end{equation}
where $\Vert$ denotes sequence concatenation. Following prior work
\cite{tang2025selfevolvingcritiqueabilitieslarge}, we balance the two
action classes through resampling, producing the balanced training set
$\mathcal D_S$. Let
$N_S=\sum_{i\in\mathcal D_S}
|\operatorname{ser}(z_i^\star)|$ be the total number of target tokens.
We then train $A_S$ with the following completion-only supervised
fine-tuning objective:
\begin{equation}
\mathcal L_S(A_S)
=
-
N_S^{-1}
\sum_{i\in\mathcal D_S}
\log M_S
\left(
z_i^\star
\mid
x_i,y_i^G
\right).
\label{eq:specialist-loss}
\end{equation}
Here, the sequence log-likelihood sums over the target tokens in
$z_i^\star$. This objective jointly trains the Specialist to decide
whether a refinement is needed and, following \texttt{APPLY}, to
generate the residual guidance.

\section{Experiments}

\subsection{Experimental Setup}

\subsubsection{Datasets and Metrics.}
We evaluate \method{} in two complementary settings: simulated feedback
on MemoryBench~\cite{ai2026memorybenchbenchmarkmemorycontinual} and real-user feedback on
WildFB~\cite{peng-etal-2026-wildreward}.

MemoryBench evaluates continual learning from
feedback logs. Following its LLM-as-user protocol, we use
Qwen3.6-Plus~\cite{qwen36plus} as the feedback simulator and generate
separate training logs from each backbone's own responses. Following the official evaluation pipeline, we evaluate held-out requests
in the four input--output length partitions and report the
sample-weighted min--max normalized score (Norm-Score) and Z-score.

WildFB contains feedback instances curated from real-world human--LLM
conversations. We randomly sample 15,000 instances from its training
split. For in-domain evaluation, we report the win rate of the adapted
model's responses over the original logged responses on the WildFB test
set, as judged by
WildReward-8B~\cite{peng-etal-2026-wildreward}. We additionally report
prompt-level strict accuracy on IFEval~\cite{zhou2023instruction},
length-controlled win rate on
AlpacaEval~2.0~\cite{dubois2025lengthcontrolledalpacaevalsimpleway}, and accuracy under
chain-of-thought prompting on
MMLU-Pro~\cite{wang2024mmlupro}.

\subsubsection{Baselines.}
We compare \method{} against the unadapted backbone (Base) and three
baseline families. They differ in how information from historical
interactions is retained and reused: through retrieved sessions in the
inference context, structured memory, or persistent parameter updates.

(\textit{i}) \emph{Direct-retrieval methods} retrieve relevant
interaction sessions as additional context at inference time, without
first converting them into structured memory. We include
BM25~\cite{robertson2009probabilistic} for sparse lexical retrieval and
Qwen3-Embedding-0.6B~\cite{zhang2025qwen3embedding} for dense semantic
retrieval.

(\textit{ii}) \emph{Memory-system methods} maintain interaction-derived
information through an explicit memory layer that is queried during
inference. A-Mem~\cite{xu2025amemagenticmemoryllm} organizes memories as an evolving
network of notes. MemoryOS~\cite{kang-etal-2025-memory} uses a hierarchy
of short-, mid-, and long-term memory, while MemOS~\cite{li2025memos}
provides unified management of heterogeneous memory representations.
ReMem~\cite{wei2025evomemory} couples task execution with continual
memory refinement.

(\textit{iii}) \emph{Parametric adaptation methods} retain
log-derived supervision through persistent parameter updates. We include
SFT~\cite{ouyang2022training} and DPO~\cite{rafailov2024directpreferenceoptimizationlanguage}.
Because interaction logs lack explicit target responses or preference
pairs, we construct chosen--rejected pairs as described in
\S~\ref{app:baseline-implementation}; SFT uses the
chosen responses, whereas DPO uses the full pairs. For
SDPO~\cite{kleinebuening2026aligning}, we use its off-policy formulation
for logged interactions, distilling the follow-up-conditioned hindsight
distribution into the policy along logged response prefixes.

We evaluate all three baseline families on MemoryBench, whose
purpose is to benchmark how LLM systems exploit accumulated
feedback through both parametric and non-parametric memory. For
models trained on WildFB, we compare Base and parametric adaptation
methods only. IFEval, AlpacaEval~2.0, and MMLU-Pro were constructed
to characterize model-level instruction following, response quality,
and knowledge-and-reasoning, respectively, while our WildReward
evaluation likewise compares response quality on held-out user
queries. Retrieval and external-memory methods would instead
evaluate a model--memory system's ability to reuse stored
interactions at inference, which constitutes a different evaluation
target.
Meanwhile, we exclude methods such as MulFeRL
\cite{li2026mulferlenhancingreinforcementlearning},
WildReward-DPO \cite{peng-etal-2026-wildreward}, and FCP
\cite{luo2025languagemodelslearnverbal}, whose protocols
respectively require task-specific verifiable rewards, reward-model
supervision from our final evaluator, and a desired feedback condition
at inference. These requirements are unavailable in our setting or
would compromise train--evaluation separation.

\subsubsection{Implementation Details.}
We conduct experiments with Qwen3-8B~\cite{yang2025qwen3} in
non-thinking mode and Ministral3-14B-Instruct~\cite{liu2026ministral3}.
For each backbone, all methods share the same training set,
context-length budget, and decoding configuration.\footnote{Further
implementation details are provided in \S~\ref{app:impl}.}
For Qwen3-8B, the extracted training components have
\textsc{Fix}/\textsc{Spec}/\textsc{Null} proportions of
23.6/50.8/25.6\% on MemoryBench and 19.4/42.9/37.7\% on WildFB.
The corresponding proportions for Ministral3-14B are
22.3/51.9/25.7\% and 18.2/43.9/38.0\%, respectively.\footnote{Role-assignment evaluation is reported in \S~\ref{app:scope-control}.}

\begin{table*}[t]
    \centering
\caption{Comparison under four input--output length partitions.
Each partition reports the sample-weighted Norm-Score and Z-score.
Results are averaged over five runs, with standard deviations shown in parentheses.
SLIFT-Gen denotes the Generalist-only variant of SLIFT.}
    \label{tab:main_results}
    \footnotesize
    \setlength{\tabcolsep}{2pt}
    \renewcommand{\arraystretch}{1.05}

    \begin{tabularx}{\textwidth}{
        @{}
        c
        c
        >{\raggedright\arraybackslash}p{0.145\textwidth}
        *{9}{>{\centering\arraybackslash}X}
        @{}
    }
        \toprule
        \multirow{2}{*}{\textbf{Model}}
        & \multirow{2}{*}{\textbf{Type}}
        & \multirow{2}{*}{\textbf{Method Name}}
        & \multicolumn{2}{c}{\textbf{Short-Short}}
        & \multicolumn{2}{c}{\textbf{Short-Long}}
        & \multicolumn{2}{c}{\textbf{Long-Short}}
        & \multicolumn{2}{c}{\textbf{Long-Long}}
        & \multicolumn{1}{c}{\textbf{Average}} \\
        \cmidrule(lr){4-5}
        \cmidrule(lr){6-7}
        \cmidrule(lr){8-9}
        \cmidrule(lr){10-11}
        \cmidrule(lr){12-12}
        &
        &
        & \mbox{\footnotesize\textbf{Norm-Score}}
        & \mbox{\footnotesize\textbf{Z-Score}}
        & \mbox{\footnotesize\textbf{Norm-Score}}
        & \mbox{\footnotesize\textbf{Z-Score}}
        & \mbox{\footnotesize\textbf{Norm-Score}}
        & \mbox{\footnotesize\textbf{Z-Score}}
        & \mbox{\footnotesize\textbf{Norm-Score}}
        & \mbox{\footnotesize\textbf{Z-Score}}
        & \mbox{\footnotesize\textbf{Norm-Score}} \\
        \midrule

        \multirow{12}{*}{\rotatebox{90}{\textbf{Qwen3-8B}}}
        & --
        & \textbf{Base}
        & \eststd{59.72}{.51} & \eststd{-1.03}{.05}
        & \eststd{53.68}{.50} & \eststd{-1.41}{.04}
        & \eststd{36.55}{.40} & \eststd{-0.48}{.03}
        & \eststd{46.74}{.46} & \eststd{-0.64}{.03}
        & \eststd{49.17}{.28} \\
        \cmidrule(lr){2-12}

        & \multirow{2}{*}{\textbf{RAG}}
        & \textbf{Embedding}
        & \eststd{59.89}{.51} & \eststd{-0.97}{.05}
        & \eststd{52.21}{.62} & \eststd{-1.48}{.05}
        & \eststd{37.44}{.51} & \eststd{-0.43}{.03}
        & \eststd{45.78}{.52} & \eststd{-0.72}{.03}
        & \eststd{48.83}{.33} \\

        &
        & \textbf{BM25}
        & \eststd{60.16}{.49} & \eststd{-0.97}{.05}
        & \eststd{53.59}{.47} & \eststd{-1.41}{.04}
        & \eststd{37.65}{.57} & \eststd{-0.43}{.04}
        & \eststd{44.41}{.53} & \eststd{-0.73}{.03}
        & \eststd{48.95}{.31} \\
        \cmidrule(lr){2-12}

        & \multirow{4}{*}{\textbf{Memory}}
        & \textbf{MemOS}
        & \eststd{53.49}{.57} & \eststd{-1.27}{.05}
        & \eststd{51.74}{.63} & \eststd{-1.51}{.04}
        & \eststd{36.98}{.59} & \eststd{-0.63}{.04}
        & \eststd{42.81}{.58} & \eststd{-0.82}{.03}
        & \eststd{46.26}{.36} \\

        &
        & \textbf{ReMem}
        & \eststd{58.05}{.71} & \eststd{-1.07}{.05}
        & \eststd{51.14}{.58} & \eststd{-1.57}{.04}
        & \eststd{33.52}{.53} & \eststd{-0.59}{.03}
        & \eststd{43.03}{.74} & \eststd{-0.83}{.04}
        & \eststd{46.44}{.39} \\

        &
        & \textbf{A-Mem}
        & \eststd{57.68}{.64} & \eststd{-1.10}{.05}
        & \eststd{50.10}{.75} & \eststd{-1.65}{.06}
        & \eststd{37.76}{.67} & \eststd{-0.47}{.04}
        & \eststd{44.31}{.59} & \eststd{-0.76}{.04}
        & \eststd{47.46}{.40} \\

        &
        & \textbf{MemoryOS}
        & \eststd{57.93}{.62} & \eststd{-1.06}{.06}
        & \eststd{54.18}{.76} & \eststd{-1.42}{.05}
        & \eststd{28.77}{.76} & \eststd{-0.76}{.05}
        & \eststd{31.16}{.80} & \eststd{-1.32}{.04}
        & \eststd{43.01}{.44} \\
        \cmidrule(lr){2-12}

        & \multirow{5}{*}{\textbf{Training}}
        & \textbf{DPO}
        & \eststd{59.62}{.54} & \eststd{-1.11}{.05}
        & \eststd{49.91}{.64} & \eststd{-1.66}{.05}
        & \eststd{35.98}{.50} & \eststd{-0.54}{.03}
        & \eststd{45.73}{.69} & \eststd{-0.71}{.03}
        & \eststd{47.81}{.36} \\

        &
        & \textbf{SFT}
        & \eststd{61.44}{.66} & \eststd{-0.99}{.06}
        & \eststd{49.25}{.71} & \eststd{-1.69}{.06}
        & \eststd{37.82}{.46} & \eststd{-0.51}{.03}
        & \eststd{46.02}{.51} & \eststd{-0.68}{.03}
        & \eststd{48.63}{.36} \\


        &
        & \textbf{SDPO}
        & \eststd{62.31}{.64} & \eststd{-0.84}{.06}
        & \eststd{55.17}{.74} & \eststd{-1.32}{.06}
        & \eststd{38.61}{.50} & \eststd{-0.47}{.03}
        & \eststd{42.14}{.54} & \eststd{-0.85}{.03}
        & \eststd{49.56}{.37} \\

        &
        & \textbf{SLIFT-Gen}
        & \eststd{65.25}{.49} & \eststd{-0.66}{.05}
        & \eststd{55.33}{.39} & \eststd{-1.26}{.03}
        & \eststd{43.20}{.49} & \eststd{-0.17}{.03}
        & \eststd{43.72}{.48} & \eststd{-0.64}{.03}
        & \eststd{51.88}{.28} \\

        &
        & \textbf{SLIFT}
        & \eststd{\textbf{70.49}}{.43} & \eststd{\textbf{-0.45}}{.03}
        & \eststd{\textbf{58.00}}{.37} & \eststd{\textbf{-1.15}}{.03}
        & \eststd{\textbf{45.22}}{.40} & \eststd{\textbf{-0.10}}{.03}
        & \eststd{\textbf{49.68}}{.45} & \eststd{\textbf{-0.50}}{.03}
        & \eststd{\textbf{55.85}}{.25} \\

        \midrule

        \multirow{12}{*}{\rotatebox{90}{\textbf{Ministral3-14B}}}
        & --
        & \textbf{Base}
        & \eststd{61.20}{.46} & \eststd{-1.17}{.04}
        & \eststd{37.86}{.46} & \eststd{-2.54}{.04}
        & \eststd{45.46}{.41} & \eststd{-0.04}{.02}
        & \eststd{54.71}{.45} & \eststd{-0.22}{.03}
        & \eststd{49.81}{.27} \\
        \cmidrule(lr){2-12}

        & \multirow{2}{*}{\textbf{RAG}}
        & \textbf{Embedding}
        & \eststd{61.46}{.63} & \eststd{-1.10}{.06}
        & \eststd{36.28}{.72} & \eststd{-2.63}{.06}
        & \eststd{46.37}{.57} & \eststd{0.01}{.03}
        & \eststd{53.58}{.74} & \eststd{-0.31}{.04}
        & \eststd{49.42}{.40} \\

        &
        & \textbf{BM25}
        & \eststd{61.73}{.53} & \eststd{-1.09}{.05}
        & \eststd{37.54}{.67} & \eststd{-2.52}{.05}
        & \eststd{46.71}{.49} & \eststd{0.03}{.03}
        & \eststd{52.26}{.68} & \eststd{-0.33}{.04}
        & \eststd{49.56}{.36} \\
        \cmidrule(lr){2-12}

        & \multirow{4}{*}{\textbf{Memory}}
        & \textbf{MemOS}
        & \eststd{55.32}{.64} & \eststd{-1.42}{.06}
        & \eststd{35.74}{.75} & \eststd{-2.66}{.05}
        & \eststd{45.91}{.56} & \eststd{-0.18}{.04}
        & \eststd{50.87}{.75} & \eststd{-0.41}{.04}
        & \eststd{46.96}{.41} \\

        &
        & \textbf{ReMem}
        & \eststd{59.58}{.55} & \eststd{-1.22}{.05}
        & \eststd{35.16}{.75} & \eststd{-2.70}{.06}
        & \eststd{42.18}{.60} & \eststd{-0.15}{.04}
        & \eststd{50.76}{.71} & \eststd{-0.43}{.04}
        & \eststd{46.92}{.39} \\

        &
        & \textbf{A-Mem}
        & \eststd{59.06}{.68} & \eststd{-1.25}{.06}
        & \eststd{34.39}{.60} & \eststd{-2.78}{.04}
        & \eststd{46.83}{.62} & \eststd{-0.03}{.04}
        & \eststd{52.12}{.74} & \eststd{-0.34}{.04}
        & \eststd{48.10}{.40} \\

        &
        & \textbf{MemoryOS}
        & \eststd{59.19}{.75} & \eststd{-1.23}{.06}
        & \eststd{38.52}{.65} & \eststd{-2.56}{.05}
        & \eststd{37.31}{.63} & \eststd{-0.32}{.04}
        & \eststd{38.72}{.79} & \eststd{-0.86}{.05}
        & \eststd{43.44}{.43} \\
        \cmidrule(lr){2-12}

        & \multirow{5}{*}{\textbf{Training}}
        & \textbf{DPO}
        & \eststd{60.98}{.57} & \eststd{-1.25}{.05}
        & \eststd{34.21}{.69} & \eststd{-2.81}{.05}
        & \eststd{44.83}{.51} & \eststd{-0.10}{.04}
        & \eststd{53.61}{.61} & \eststd{-0.30}{.04}
        & \eststd{48.41}{.36} \\

        &
        & \textbf{SFT}
        & \eststd{63.04}{.49} & \eststd{-1.11}{.05}
        & \eststd{33.23}{.52} & \eststd{-2.85}{.04}
        & \eststd{46.79}{.62} & \eststd{-0.08}{.04}
        & \eststd{53.96}{.76} & \eststd{-0.27}{.04}
        & \eststd{49.26}{.36} \\


        &
        & \textbf{SDPO}
        & \eststd{63.40}{.61} & \eststd{-1.01}{.05}
        & \eststd{38.03}{.53} & \eststd{-2.52}{.04}
        & \eststd{43.50}{.65} & \eststd{-0.14}{.04}
        & \eststd{49.82}{.67} & \eststd{-0.41}{.03}
        & \eststd{48.69}{.37} \\

        &
        & \textbf{SLIFT-Gen}
        & \eststd{64.23}{.51} & \eststd{-0.76}{.05}
        & \eststd{39.20}{.59} & \eststd{-2.37}{.05}
        & \eststd{46.61}{.47} & \eststd{0.01}{.03}
        & \eststd{55.62}{.54} & \eststd{-0.18}{.03}
        & \eststd{51.42}{.32} \\

        &
        & \textbf{SLIFT}
        & \eststd{\textbf{67.81}}{.45} & \eststd{\textbf{-0.57}}{.04}
        & \eststd{\textbf{39.22}}{.58} & \eststd{\textbf{-2.36}}{.04}
        & \eststd{\textbf{48.75}}{.53} & \eststd{\textbf{0.10}}{.03}
        & \eststd{\textbf{57.09}}{.47} & \eststd{\textbf{-0.12}}{.03}
        & \eststd{\textbf{53.22}}{.31} \\
        \bottomrule
    \end{tabularx}
\end{table*}

\begin{figure*}[t]
\centering
\includegraphics[width=\textwidth]{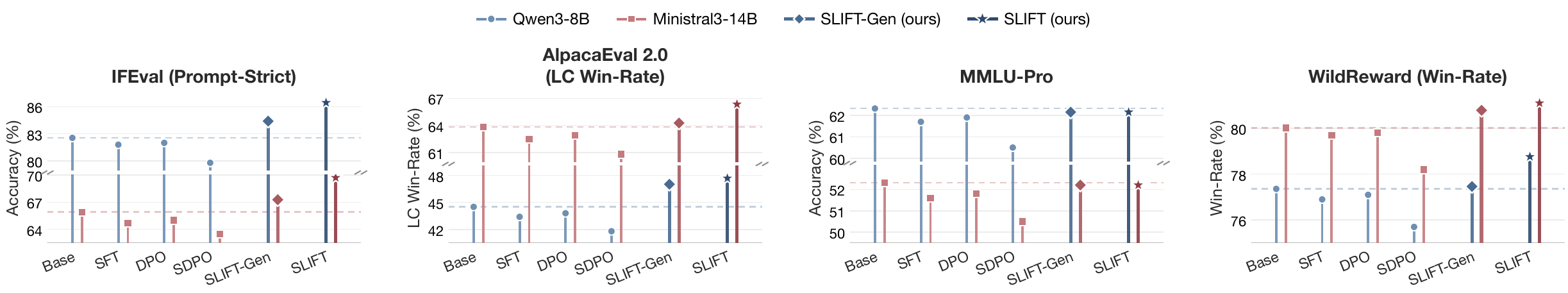}
\caption{Results on WildFB across four metrics and two backbones.}
\label{fig:wildfb}
\end{figure*}

\subsection{Main Results}
\label{exp:main}

\begingroup
\makeatletter
\renewcommand{\paragraph}{%
  \@startsection{paragraph}{4}{2em}%
    {0.3\baselineskip}
    {-0.6em}
    {\normalfont\itshape}%
}
\makeatother

\paragraph{Results after MemoryBench training.}
As shown in Table~\ref{tab:main_results}, \method{} achieves the
highest mean Norm-Score and Z-score on all four input--output length
partitions for both backbones. The gains are therefore not confined to
either tested backbone or to a single length regime. Their consistency
on held-out requests indicates that \method{} can recover useful
supervision from backbone-specific interaction trajectories and convert
it into improvements that remain available on subsequent tasks. These
results establish the overall effectiveness of the complete framework;
the contributions of its main components are examined in the ablation
study.

\paragraph{Results after WildFB training.}
As shown in Figure~\ref{fig:wildfb}, \method{}
improves IFEval and AlpacaEval~2.0 for both backbones, yields a modest
positive gain in WildReward win rate on held-out WildFB queries, and
maintains comparable performance on MMLU-Pro. The first two gains
indicate that supervision from real interaction logs transfers to
held-out evaluations of instruction adherence and open-ended response
quality. The stable MMLU-Pro results further suggest that these
behavioral gains do not come with a material loss on the evaluated
knowledge-and-reasoning benchmark.

Because WildFB is derived from
WildChat~\cite{zhao2024wildchat1mchatgptinteraction}, its logged
assistant responses were produced by GPT-family models rather than the
evaluated backbones. WildReward thus provides a cross-policy comparison
against an external logged policy. Even in this setting, \method{}
modestly improves the win rate over Base. As analyzed below, the larger
share of WildFB components without reliable update targets may partly
account for the limited in-domain gain. Overall, \method{} converts real
interaction feedback into persistent behavioral gains while maintaining
comparable MMLU-Pro performance.

\paragraph{Comparison with Other Methods.}
Non-parametric baselines yield smaller consistent gains on MemoryBench, indicating that retaining interaction-derived information outside the model does not by itself ensure transfer to held-out requests. Across both settings, no parametric baseline matches
\method{} consistently. These methods construct supervision from entire-response targets, preference pairs, or follow-up-conditioned distributions. In contrast, \method{} decomposes feedback into atomic components and distinguishes task-necessary corrections, conditional refinements, and components
without reliable update targets before adaptation. This performance pattern is consistent with our central hypothesis that composite feedback should not be treated as a single undifferentiated supervision unit. The following ablation study further tests this explanation.

\endgroup

\subsection{Ablation Study}
\label{sec:ablation}

\begin{table}[t]
    \centering
    \caption{Ablation study on Ministral3-14B. The MemoryBench
    column uses models trained on MemoryBench logs, and the remaining
    columns use models trained on WildFB logs. We report means over
    five runs, with standard deviations in parentheses.}
    \label{tab:ablation-slift}
    \footnotesize
    \setlength{\tabcolsep}{1.0pt}
    \renewcommand{\arraystretch}{1.08}

    \begin{tabularx}{\columnwidth}{
        @{}
        >{\raggedright\arraybackslash}X
        >{\centering\arraybackslash}p{0.18\columnwidth}
        >{\centering\arraybackslash}p{0.18\columnwidth}
        >{\centering\arraybackslash}p{0.20\columnwidth}
        @{}
    }
        \toprule
        \textbf{Method}
        &
        \textbf{MemoryBench}\par
        \vspace{-0.25ex}
        {\scriptsize\textbf{Avg. Norm-Score}}
        &
        \textbf{IFEval}\par
        \vspace{-0.25ex}
        {\scriptsize\textbf{Prompt-Strict}}
        &
        \textbf{AlpacaEval 2.0}\par
        \vspace{-0.25ex}
        {\scriptsize\textbf{LC Win Rate}} \\
        \midrule

        \textbf{\method{}}
        & \eststd{\textbf{53.22}}{.31}
        & \eststd{\textbf{69.71}}{1.02}
        & \eststd{\textbf{66.35}}{.21} \\

        \midrule

        w/o Atomic Extraction
        & \eststd{51.30}{.34}
        & \eststd{67.89}{.91}
        & \eststd{65.22}{.19} \\

        w/o Task Context for Roles
        & \eststd{50.74}{.36}
        & \eststd{67.34}{.74}
        & \eststd{63.98}{.22} \\

        w/o Separate Pathways
        & \eststd{51.05}{.33}
        & \eststd{67.62}{.86}
        & \eststd{64.61}{.18} \\

        w/o Specialist (\method{}-Gen)
        & \eststd{51.42}{.32}
        & \eststd{67.28}{.79}
        & \eststd{64.27}{.17} \\

        w/o \texttt{KEEP} Supervision
        & \eststd{50.86}{.44}
        & \eststd{57.63}{1.36}
        & \eststd{64.06}{.28} \\

        \bottomrule
    \end{tabularx}
\end{table}

Table~\ref{tab:ablation-slift} evaluates five central design choices
in \method{}. \emph{w/o Atomic Extraction} assigns a single role to
the complete feedback turn, while \emph{w/o Task Context for Roles}
removes the original task from role assignment. \emph{w/o Separate
Pathways} sends both \textsc{Fix} and \textsc{Spec} through the
Generalist, while continuing to exclude \textsc{Null}.
\emph{w/o Specialist}, corresponding to \method{}-Gen, discards
\textsc{Spec} supervision and retains only the Generalist.
\emph{w/o \texttt{KEEP} Supervision} trains the Specialist only on
\texttt{APPLY} targets.
Removing atomic extraction lowers performance across all three
evaluations, while assigning roles without the original task produces
the largest drops on MemoryBench and AlpacaEval~2.0. Composite
feedback must therefore be separated into atomic signals, and each
signal must be interpreted relative to the task when distinguishing
\textsc{Fix}, \textsc{Spec}, and \textsc{Null}.
The pathway variants show that \textsc{Spec} should neither be
consolidated through the default Generalist pathway nor discarded.
Both alternatives underperform the complete model, supporting a
separate pathway for response-conditioned refinement. Removing
\texttt{KEEP} supervision is especially harmful on IFEval: without
explicit supervision for when not to intervene, the Specialist risks
over-applying refinements and disrupting already satisfied instruction
constraints.

\section{Analysis}

\begin{figure}[t]
    \centering

    \begin{subfigure}[t]{0.48\linewidth}
        \centering
        \includegraphics[width=\linewidth]
            {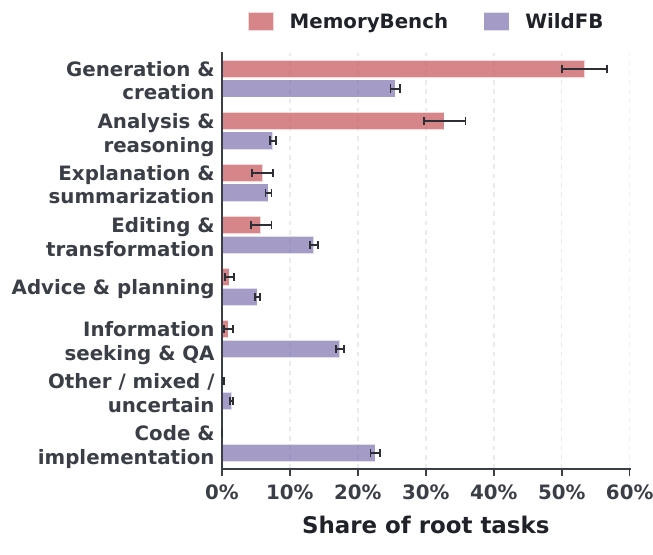}
        \caption{Task-operation distribution.}
        \label{fig:task-composition}
    \end{subfigure}
    \hfill
    \begin{subfigure}[t]{0.48\linewidth}
        \centering
        \includegraphics[width=\linewidth]
            {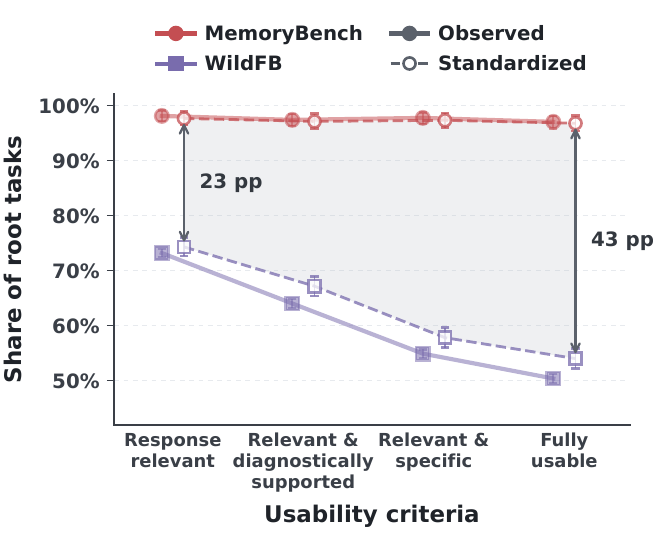}
        \caption{Feedback Usability.}
        \label{fig:feedback-yields}
    \end{subfigure}

    \caption{Task composition and feedback usability in the MemoryBench
    and WildFB training logs. Standardized rates match the two logs over
    shared task-operation and verifiability strata.}
    \label{fig:task-feedback-usability}
\end{figure}

\subsection{Task Composition and Feedback Usability}
\label{sec:task-feedback-usability}

We analyze the first feedback opportunity for each root training task.
A Qwen3.7-Max annotator \cite{qwen37} labels the task operation from
the task context and evaluates the feedback using the logged response
along four dimensions: response relevance, diagnostic support,
specificity, and actionability. Feedback satisfying all four criteria
is considered fully usable. For the standardized estimates, we stratify
tasks by operation and verifiability and reweight both training logs to
the same equal-source distribution over shared strata.
\footnote{Full annotation and standardization details are provided in
\S~\ref{app:task-feedback-usability}.}

\textit{The training logs cover different task distributions.}
Figure~\ref{fig:task-composition} shows that MemoryBench is concentrated
in generation and creation (53.4\%) and analysis and reasoning (32.7\%).
Its training logs and test tasks are drawn from separate splits of the
same underlying datasets, yielding close alignment between the training
and test task distributions. WildFB instead
contains larger shares of code and implementation (22.6\%), information
seeking and question answering (17.4\%), and editing and transformation
(13.6\%). This mixture is better aligned with the user-facing assistant
behaviors evaluated by IFEval and AlpacaEval 2.0, but provides little direct
coverage of the knowledge-intensive reasoning tasks in
MMLU-Pro. The task alignment therefore helps
explain why training on MemoryBench transfers strongly to its test
partitions, while training on WildFB improves IFEval and AlpacaEval 2.0
but leaves MMLU-Pro largely unchanged.

\textit{Simulated feedback is denser and more usable.}
As shown in Figure~\ref{fig:feedback-yields}, fully usable feedback is
present in 97.0\% of MemoryBench root tasks, compared with 50.4\% of
WildFB root tasks. After matching task operation and verifiability, the
rates remain 96.7\% and 54.0\%, respectively. In these training logs,
simulated users therefore provide denser and more directly usable
supervision than naturally occurring real-user feedback. Together, task coverage shapes where the learned changes transfer, while feedback usability shapes the strength of that transfer.

\subsection{Stage-wise Contributions}
\label{sec:stage-wise-transfer}

We compare Base, \method{}-Gen, and the complete \method{} to separate
the gains introduced by the Generalist and Specialist.

\begin{table}[t]
\centering
\footnotesize
\renewcommand{\arraystretch}{1.06}
\caption{Stage-wise gains on Ministral3-14B.
$\Delta_{\mathrm{G}}$ and $\Delta_{\mathrm{S}}$ denote the absolute
point changes from Base to \method{}-Gen and from \method{}-Gen to
\method{} on each evaluation's main metric.}
\label{tab:stage-transfer}

\begin{tabular*}{\columnwidth}{
    @{\extracolsep{\fill}}
    l
    c
    c
    c
    c
    c
    @{}
}
\toprule
\textbf{Eval.}
& \textbf{\#Tasks}
& \textbf{$\Delta_{\mathrm{G}}$}
& \textbf{\texttt{APPLY} (\%)}
& \textbf{\texttt{Edit} (\%)}
& \textbf{$\Delta_{\mathrm{S}}$} \\
\midrule

\multicolumn{6}{@{}l}{\textbf{\textit{Trained on MemoryBench}}} \\
Short--Short
& 150
& $+3.03$
& 19.33
& 18.00
& $+3.58$ \\

Short--Long
& 143
& $+1.34$
& 1.40
& 1.40
& $+0.02$ \\

Long--Short
& 150
& $+1.15$
& 10.67
& 10.00
& $+2.14$ \\

Long--Long
& 275
& $+0.91$
& 9.09
& 8.36
& $+1.47$ \\

\midrule

\multicolumn{6}{@{}l}{\textbf{\textit{Trained on WildFB}}} \\
IFEval
& 541
& $+1.36$
& 14.23
& 13.12
& $+2.43$ \\

AlpacaEval 2.0
& 805
& $+0.45$
& 12.80
& 11.80
& $+2.08$ \\

MMLU-Pro
& 12{,}032
& $-0.11$
& 0.00
& 0.00
& $+0.00$ \\

WildReward
& 4{,}929
& $+0.77$
& 2.50
& 2.19
& $+0.31$ \\
\bottomrule
\end{tabular*}

\vspace{0.3em}
\parbox{\columnwidth}{\footnotesize
\textit{Note:} \#Tasks is the number of test tasks.
\texttt{APPLY} is the Specialist's apply-decision rate, and \texttt{Edit} is
the rate at which the final response differs from the Generalist
response.}
\end{table}

\textit{The Generalist provides broad gains.}
Training the Generalist on MemoryBench improves all four test
partitions. For models trained on WildFB, it improves IFEval,
AlpacaEval~2.0, and WildReward, while MMLU-Pro remains nearly unchanged.
Thus, consolidating \textsc{Fix} components transfers improvements from
the training logs to unseen test tasks.

\textit{The Specialist adds targeted gains with sparse edits.}
Across MemoryBench, the Specialist improves the macro-average and
benefits three of the four length partitions, while leaving
Short--Long unchanged. For models trained on WildFB logs, it further
improves IFEval, AlpacaEval~2.0, and WildReward, but has no additional
effect on MMLU-Pro. Edit rates remain below $20\%$ across all test
sets, with the Specialist nearly inactive on Short--Long and entirely
inactive on MMLU-Pro. These results indicate that the Specialist
contributes through sparse, task-dependent refinements rather than
uniform editing.

\subsection{Online Evolution}
\label{sec:online-training}

\begin{figure}[t]
  \begin{center}
    \centerline{\includegraphics[width=0.8\columnwidth]
    {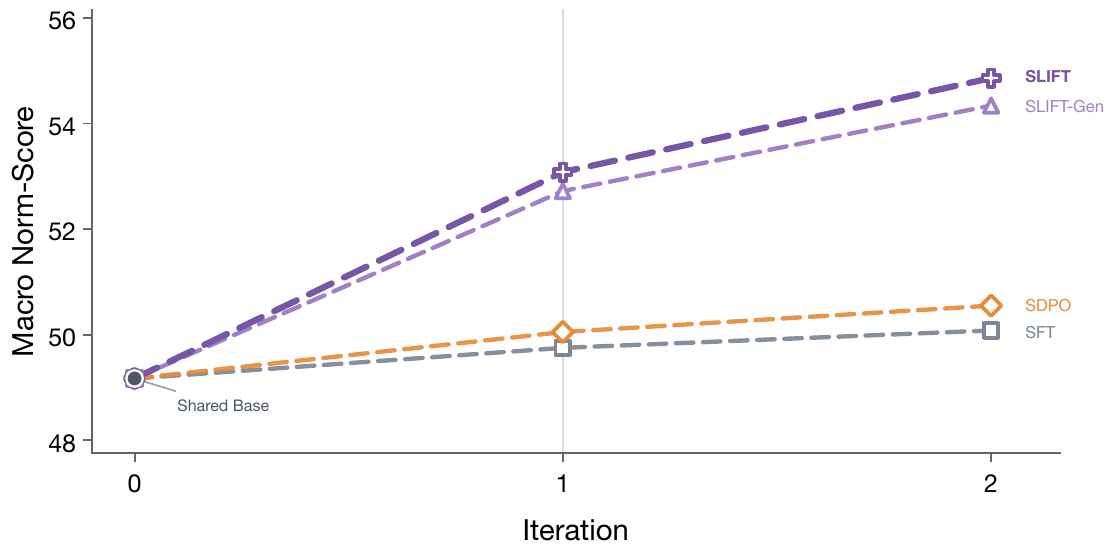}}
    \caption{Online evolution on MemoryBench (Qwen3-8B).}
    \label{fig:online_evolution}
  \end{center}
  \vspace{-2.0em}
\end{figure}

We evenly split the MemoryBench training set into two disjoint
batches. In the first iteration, each method is trained on the
pre-collected logs from the first batch. The resulting model is then
deployed on the second-batch tasks, where the MemoryBench User
Simulator generates feedback conditioned on its responses. These
newly collected on-policy logs are used for a second incremental
update. We evaluate all models after each iteration on the same fixed
MemoryBench test set.
As shown in Figure~\ref{fig:online_evolution}, all methods improve
after the first training round and continue to benefit from the
on-policy logs collected in the second round. \method{} achieves the
highest Norm-Score at both iterations and further increases its
lead over the baselines after the second update. \method{}-Gen follows
the same trend but remains below the complete model, showing that the
Specialist contributes complementary gains throughout sequential
training. Overall, \method{} extracts greater gains from both
pre-collected and on-policy feedback, indicating that its selective
updates remain effective as the training logs shift toward interactions
generated by the current model.

\begin{figure}[t]
\centering
\includegraphics[width=\columnwidth]{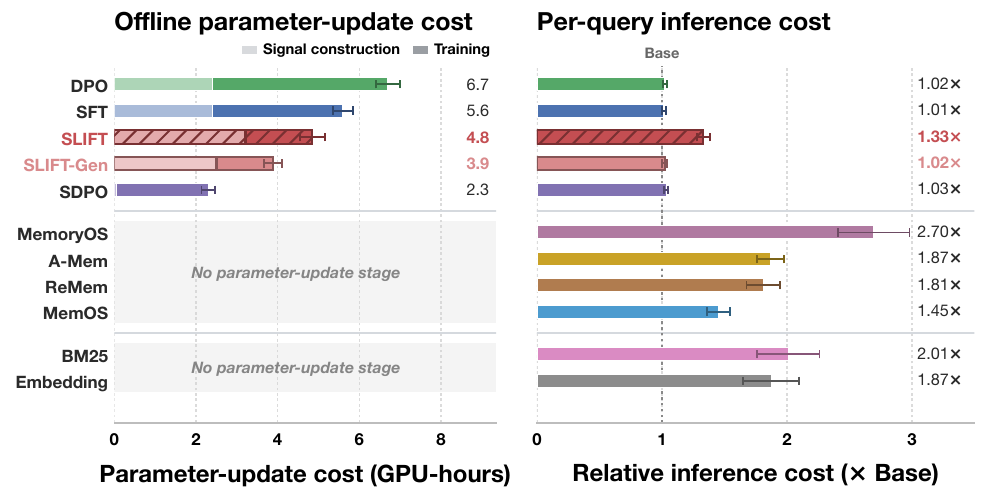}
\caption{Offline parameter-update and per-query inference costs on Qwen3-8B.
Parameter-update cost includes signal construction and model training,
while inference cost is measured per query and normalized to Base.}
\label{fig:efficiency}
\end{figure}

\subsection{Efficiency}
\label{ana:efficiency}

Figure~\ref{fig:efficiency} compares the offline parameter-update and
per-query inference costs of different methods. \method{} incurs
additional preprocessing for feedback decomposition and Specialist
target construction, but avoids generating full revised targets. The
Generalist distills \textsc{Fix}-conditioned distributions, while the
Specialist learns to output either \texttt{KEEP}, with no residual
guidance, or \texttt{APPLY} together with compact residual guidance.
This keeps the total parameter-update cost below SFT and DPO,
although SDPO remains cheaper. The small gap between \method{}-Gen and
\method{} shows that Specialist training adds limited overhead.
At inference time, \method{} does not retrieve the training logs. After
the Generalist produces a response, the Specialist performs a single
constrained generation over the two output forms. A \texttt{KEEP}
output contains no guidance and returns the Generalist response
directly, whereas an \texttt{APPLY} output includes residual guidance
and triggers one integration pass. The low \texttt{APPLY} and edit
rates in Table~\ref{tab:stage-transfer} show that this integration pass
is required for only a small subset of test tasks, allowing the
Specialist to add targeted gains with limited average overhead. Its
average inference cost is therefore higher than that of single-pass
SFT, DPO, and SDPO, but lower than that of the retrieval and
external-memory methods. Because the learned changes are stored in
parameters, its per-query cost does not grow with the size of the
training log.

\section{Conclusion}
\label{sec:conclusion}

We introduced \method{}, a selective self-learning framework that converts
interaction feedback into persistent behavioral improvements through a
task-relative view. It decomposes feedback into atomic \textsc{Fix},
\textsc{Spec}, and \textsc{Null} components, corresponding to task-necessary
requirements, compatible refinements, and no reliable positive update. The
Generalist consolidates \textsc{Fix}-derived requirements into default
behavior, while the Specialist uses the task and Generalist response to supply
residual guidance for applicable, unmet \textsc{Spec} refinements. Both
pathways use supervision from the frozen backbone, without a stronger teacher
or explicit reward model. Across backbones, \method{} achieves strong performance on both MemoryBench and WildFB, highlighting the importance of learning both the supported behavioral change and its appropriate scope.

\bibliographystyle{ACM-Reference-Format}
\bibliography{sample-base}

\appendix

\section{Algorithmic Details of \method{}}
\label{app:algorithmic-details}

This section provides the complete offline training and inference
procedures for \method{}. All feedback-processing and target-construction
operations are performed by the frozen backbone $M_0$. During
optimization, only the Generalist adapter $A_G$ or the Specialist
adapter $A_S$ is updated at its corresponding stage.

\subsection{Offline Feedback Processing}
\label{app:offline-processing}

For each logged interaction
$d_i=(x_i,y_i^{\mathrm{log}},u_i)$, the frozen backbone first extracts
an ordered sequence of atomic components:
\begin{equation}
\mathcal C_i
=
\operatorname{Extract}_{M_0}
\left(
x_i,
y_i^{\mathrm{log}},
u_i
\right)
=
\left(
c_{i1},
\ldots,
c_{im_i}
\right).
\label{eq:app-extract}
\end{equation}
Each component is then assigned a task-relative role:
\begin{equation}
r_{ij}
=
\operatorname{Assign}_{M_0}
\left(
x_i,c_{ij}
\right)
\in
\left\{
\mathrm{FIX},
\mathrm{SPEC},
\mathrm{NULL}
\right\}.
\label{eq:app-assign}
\end{equation}
The assignment prompt implements the actionability, compatibility,
and omission criteria described in
\S~\ref{sec:projection}.

We serialize the components assigned to each positive role while
preserving their original order and boundaries:
\begin{align}
C_i^F
&=
\operatorname{Serialize}
\left(
\left(
c_{ij}
\right)_{j:r_{ij}=\mathrm{FIX}}
\right),
\label{eq:app-fix-serialize}
\\
C_i^S
&=
\operatorname{Serialize}
\left(
\left(
c_{ij}
\right)_{j:r_{ij}=\mathrm{SPEC}}
\right).
\label{eq:app-spec-serialize}
\end{align}
Either sequence may be empty. The Generalist training set is defined as:
\begin{equation}
\mathcal D_F
=
\left\{
d_i\in\mathcal D
\mid
C_i^F\neq\varnothing
\right\}.
\label{eq:app-generalist-set}
\end{equation}

\subsection{Generalist Optimization}
\label{app:generalist-algorithm}

Let $M_G=M_0\oplus A_G$, where $A_G$ is the trainable Generalist
adapter. For interaction $i$ and logged position $t$, define:
\begin{align}
p_{i,t}^{G}
&=
M_G
\left(
\cdot
\mid
x_i,y_{i,<t}^{\mathrm{log}}
\right),
\label{eq:app-pg}
\\
p_{i,t}^{H}
&=
M_0
\left(
\cdot
\mid
x_i,C_i^F,y_{i,<t}^{\mathrm{log}}
\right),
\label{eq:app-ph}
\\
p_{i,t}^{B}
&=
M_0
\left(
\cdot
\mid
x_i,y_{i,<t}^{\mathrm{log}}
\right).
\label{eq:app-pb}
\end{align}
Both $p_{i,t}^{H}$ and $p_{i,t}^{B}$ are detached because $M_0$ is
frozen. The Generalist is optimized by teacher forcing over the
logged prefixes:
\begin{equation}
\begin{split}
\mathcal L_G(A_G)
&=
\mathbb E_{d_i\sim\mathcal D_F}
\frac{1}{T_i}
\sum_{t=1}^{T_i}
\left[
D_{\mathrm{KL}}
\left(
p_{i,t}^{G}
\middle\|
p_{i,t}^{H}
\right)
\right.
\\[-1mm]
&\qquad\left.
+
\lambda_B
D_{\mathrm{KL}}
\left(
p_{i,t}^{G}
\middle\|
p_{i,t}^{B}
\right)
\right],
\qquad
T_i=\left|y_i^{\mathrm{log}}\right|.
\end{split}
\label{eq:app-generalist-loss}
\end{equation}
The response-level normalization prevents longer logged responses
from receiving proportionally larger weight.

After optimizing $A_G$, we freeze the resulting Generalist $M_G$.
For every processed interaction, including those without a
\textsc{Fix}, we then generate:
\begin{equation}
y_i^G
\sim
M_G
\left(
\cdot
\mid
x_i
\right),
\label{eq:app-generalist-generation}
\end{equation}
using the fixed decoding configuration specified in the experimental
setup.

\subsection{Specialist Target Construction}
\label{app:specialist-targets}

After Generalist training, we freeze $M_G$ and generate a default
response $y_i^G$ for each processed training interaction. Let $C_i^S$
denote the possibly empty ordered sequence of \textsc{Spec} components
assigned in \S~\ref{sec:projection}. During offline target
construction, $C_i^S$ is provided to the frozen backbone as privileged
information, while the Specialist uses only $(x_i,y_i^G)$ as input.

Let $\Delta^+$ denote the set of nonempty residual-guidance sequences.
The structured target space is
\begin{equation}
\mathcal T_S
=
\left\{
\left(
\texttt{KEEP},
\varnothing
\right)
\right\}
\cup
\left\{
\left(
\texttt{APPLY},
\delta
\right)
\mid
\delta\in\Delta^+
\right\}.
\label{eq:app-specialist-output-space}
\end{equation}
A \texttt{KEEP} target is paired with an empty guidance field, whereas
an \texttt{APPLY} target contains nonempty residual guidance.

For $C_i^S\neq\varnothing$, the frozen backbone constructs both fields
in one structured call:
\begin{equation}
\left(
a_i^\star,
\delta_i^\star
\right)
=
\operatorname{Target}_{M_0}
\left(
x_i,
y_i^G,
C_i^S
\right)
\in
\mathcal T_S.
\label{eq:app-specialist-target}
\end{equation}
It returns
$\left(\texttt{KEEP},\varnothing\right)$ when no supported
\textsc{Spec} refinement both applies to the current task and remains
unmet in $y_i^G$. Otherwise, it returns
$\left(\texttt{APPLY},\delta_i^\star\right)$, where
$\delta_i^\star$ describes only the compatible refinements that remain
unmet in $y_i^G$.

When $C_i^S=\varnothing$, we directly set
\begin{equation}
\left(
a_i^\star,
\delta_i^\star
\right)
=
\left(
\texttt{KEEP},
\varnothing
\right).
\label{eq:app-empty-spec-target}
\end{equation}

For training, we serialize
$\left(a_i^\star,\delta_i^\star\right)$ as a single completion
$z_i^\star$: a \texttt{KEEP} action is serialized as \texttt{KEEP},
whereas an \texttt{APPLY} action is serialized as \texttt{APPLY}
followed by $\delta_i^\star$. The resulting candidate pool is
\begin{equation}
\widetilde{\mathcal D}_S
=
\left\{
\left(
x_i,
y_i^G,
z_i^\star
\right)
\right\}_{i=1}^{N}.
\label{eq:app-specialist-candidates}
\end{equation}
Following the 1:1 balancing used in prior LLM critic training
\cite{tang2025selfevolvingcritiqueabilitieslarge}, we upsample
\texttt{KEEP} candidates to match the number of \texttt{APPLY}
candidates, yielding the balanced training set $\mathcal D_S$.

\subsection{Specialist Optimization}
\label{app:specialist-algorithm}

We obtain $M_S=M_0\oplus A_S$ by attaching a trainable LoRA adapter
$A_S$ to the frozen backbone, initialized independently of $A_G$.
Each example uses $(x_i,y_i^G)$ as the prompt and $z_i^\star$ as the
completion. Let
$T_i=|z_i^\star|$ and
$N_S=\sum_{i\in\mathcal D_S}T_i$ denote the completion length and the
total number of target tokens, respectively. We optimize:
\begin{equation}
\mathcal L_S(A_S)
=
-
N_S^{-1}
\sum_{i\in\mathcal D_S}
\sum_{t=1}^{T_i}
\log
M_S
\left(
z_{i,t}^\star
\mid
x_i,
y_i^G,
z_{i,<t}^\star
\right).
\label{eq:app-specialist-loss}
\end{equation}
The loss is computed over the completion tokens in $z_i^\star$. It
jointly trains the Specialist to select \texttt{KEEP} or
\texttt{APPLY} and, following \texttt{APPLY}, to generate the residual
guidance.

At inference time, the Specialist receives $(x_i,y_i^G)$ and generates
one output constrained to $\mathcal T_S$. A \texttt{KEEP} output
returns $y_i^G$ unchanged. An \texttt{APPLY} output provides nonempty
residual guidance $\delta_i$, which the frozen Generalist incorporates
through a single residual-guided integration pass.

\subsection{Inference}
\label{app:inference-algorithm}

At inference time, the two adapters and the backbone remain frozen.
The Generalist first generates its default response:
\begin{equation}
y^G
=
\operatorname{Decode}_{M_G}(x).
\label{eq:app-inference-generalist}
\end{equation}
Using only $(x,y^G)$, the Specialist then performs one structured
decoding call constrained to the target space defined in
Equation~\eqref{eq:app-specialist-output-space}:
\begin{equation}
\left(
a,
\delta
\right)
=
\operatorname{StructuredDecode}_{M_S}
\left(
\mathcal T_S
\mid
x,
y^G
\right)
\in
\mathcal T_S.
\label{eq:app-inference-specialist}
\end{equation}
The constraint is applied to the complete structured output rather
than to an isolated action. Consequently, the single Specialist output
is either $(\texttt{KEEP},\varnothing)$ or
$(\texttt{APPLY},\delta)$ with $\delta\in\Delta^+$.

If $a=\texttt{KEEP}$, the structured-output constraint guarantees
$\delta=\varnothing$, and the Generalist response is returned
unchanged. If $a=\texttt{APPLY}$, the same output contains nonempty
residual guidance $\delta$, and the frozen Generalist performs one
integration pass:
\begin{equation}
y^{\mathrm{final}}
=
\operatorname{Integrate}_{M_G}
\left(
x,
y^G,
\delta
\right).
\label{eq:app-residual-integration}
\end{equation}
The fixed integration instruction asks the model to implement only the
changes specified by $\delta$ and preserve unaffected parts of $y^G$.
No second Specialist generation or iterative editing is performed.

\begin{algorithm}[t]
\caption{\method{} inference}
\label{alg:slift-inference}
\begin{algorithmic}[1]
\Require User task $x$; frozen $M_G$ and $M_S$
\Ensure Final response $y^{\mathrm{final}}$

\State
$y^G
\gets
\operatorname{Decode}_{M_G}(x)$

\State
$(a,\delta)
\gets
\operatorname{StructuredDecode}_{M_S}
(\mathcal T_S\mid x,y^G)$

\If{$a=\texttt{KEEP}$}
    \State $y^{\mathrm{final}} \gets y^G$
\Else
    \State
    $y^{\mathrm{final}}
    \gets
    \operatorname{Integrate}_{M_G}
    (x,y^G,\delta)$
\EndIf
\State \Return $y^{\mathrm{final}}$
\end{algorithmic}
\end{algorithm}

\subsection{Logged-Prefix Self-Distillation}
\label{app:offline-sdpo}

The Generalist is trained on prefixes of logged responses, which need
not have been generated by the current Generalist. We therefore make
explicit the relation between Equation~\eqref{eq:app-generalist-loss}
and the corresponding on-policy objective.

Let $\theta$ denote the parameters of $A_G$ and write
$\pi_\theta=M_0\oplus A_G$. For interaction $i$ and an arbitrary
response prefix $h$, define the fixed target distributions:
\begin{align}
q_i^H(\cdot\mid h)
&=
M_0
\left(
\cdot
\mid
x_i,
C_i^F,
h
\right),
\\
q_i^B(\cdot\mid h)
&=
M_0
\left(
\cdot
\mid
x_i,
h
\right),
\end{align}
We also define the per-prefix loss:
\begin{equation}
\begin{split}
\ell_\theta(i,h)
&=
D_{\mathrm{KL}}
\left(
\pi_\theta(\cdot\mid x_i,h)
\middle\|
q_i^H(\cdot\mid h)
\right)
\\
&\quad
+
\lambda_B
D_{\mathrm{KL}}
\left(
\pi_\theta(\cdot\mid x_i,h)
\middle\|
q_i^B(\cdot\mid h)
\right).
\end{split}
\label{eq:app-prefix-loss}
\end{equation}

Let $\nu_{\mathrm{log}}$ be the distribution obtained by sampling
$d_i\sim\mathcal D_F$, sampling a token position uniformly from its
logged response, and returning the corresponding prefix. The resulting
logged-prefix objective is:
\begin{equation}
\mathcal L_G(\theta)
=
\mathbb E_{(I,H)\sim\nu_{\mathrm{log}}}
\left[
\ell_\theta(I,H)
\right],
\label{eq:app-logged-surrogate}
\end{equation}
which is the off-policy logged-interaction surrogate of SDPO
\cite{kleinebuening2026aligning}, specialized to \textsc{Fix}
context and augmented with the behavioral anchor.

For comparison, at optimization iterate $\theta_k$, an on-policy
version would sample responses from $\pi_{\theta_k}$ and induce a
prefix distribution $\nu_{\theta_k}$. With the rollout distribution
detached during the update, its gradient is given by:
\begin{equation}
g_{\mathrm{on}}(\theta_k)
=
\left.
\mathbb E_{(I,H)\sim\nu_{\theta_k}}
\left[
\nabla_\theta\ell_\theta(I,H)
\right]
\right|_{\theta=\theta_k},
\end{equation}
whereas the logged-prefix update is given by:
\begin{equation}
g_{\mathrm{log}}(\theta_k)
=
\left.
\mathbb E_{(I,H)\sim\nu_{\mathrm{log}}}
\left[
\nabla_\theta\ell_\theta(I,H)
\right]
\right|_{\theta=\theta_k}.
\end{equation}
If
$\|\nabla_\theta\ell_\theta(i,h)|_{\theta=\theta_k}\|\leq G$
on the support of both prefix distributions, then the following bound
holds:
\begin{equation}
\left\|
g_{\mathrm{log}}(\theta_k)
-
g_{\mathrm{on}}(\theta_k)
\right\|
\leq
2G\,
D_{\mathrm{TV}}
\left(
\nu_{\mathrm{log}},
\nu_{\theta_k}
\right).
\label{eq:app-off-policy-bound}
\end{equation}
This follows directly by writing the difference as an integral against
$d\nu_{\mathrm{log}}-d\nu_{\theta_k}$ and applying the triangle
inequality. Exact correction would require logging-policy or
prefix-density ratios, which are unavailable for the interaction logs.
Accordingly, \method{} optimizes the unweighted surrogate in
Equation~\eqref{eq:app-logged-surrogate}. The behavioral anchor
regularizes changes at observed prefixes.

\section{Implementation Details}
\label{app:impl}

\subsection{Backbones and Data Construction}

We use the instruction-tuned checkpoints Qwen3-8B and
Ministral3-14B-Instruct as $M_0$. Qwen3-8B is run with its official
non-thinking chat template. For a given backbone, the same frozen
checkpoint is used for atomic-component extraction, task-relative role
assignment, the two Generalist teacher distributions, and offline
Specialist-target construction. Thus, none of these stages introduces
a stronger teacher model. The Generalist and Specialist LoRA adapters
are initialized independently from $M_0$; they are never stacked in a
single forward pass.

For MemoryBench, we generate a separate interaction log from each
backbone's own responses, while Qwen3.6-Plus is used only as the
user simulator under the benchmark's interaction protocol. For
WildFB, we use the 15,000 training interactions sampled as described in
the main text. Dataset identifiers, satisfaction labels, evaluation
metadata, and test-set information are excluded from all prompts and
training targets. All offline structured generations used to construct
\method{} supervision use deterministic decoding. Baseline-specific
stochastic operations follow their original protocols, as described in
\S~\ref{app:baseline-implementation}.

\subsection{Prompt Templates}
\label{app:prompts}

\tcbset{promptbox/.append style={
  colframe=blue!45!black,
  colbacktitle=blue!45!black
}}

The bracketed fields below are replaced verbatim and wrapped with the
native chat template of the corresponding backbone. All structured
calls request JSON only and do not request chain-of-thought.
Extraction observes the complete logged interaction, whereas role
assignment observes only the original task and one extracted
component, matching the information boundaries defined in the main
method.

\paragraph{Atomic-component extraction.}
The extractor may return an empty list. A separate new request is still
extracted as a component and is handled by role assignment.

\begin{figure}[t]
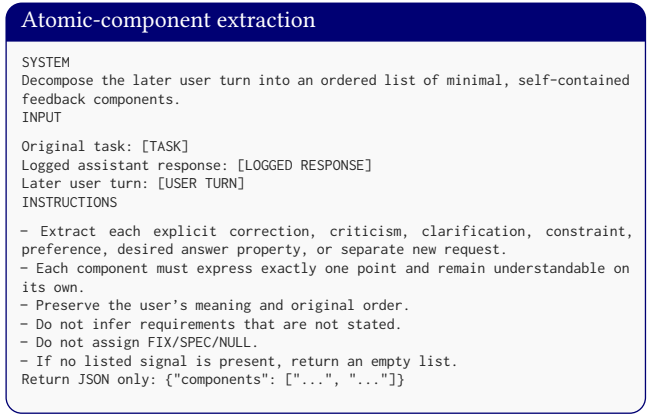

\centering
\begin{tcolorbox}[
  promptbox,
  title=Atomic-component extraction
]
\scriptsize\ttfamily
\color{black!85}
SYSTEM\\
Decompose the later user turn into an ordered list of minimal,
self-contained feedback components.\\
\medskip
INPUT\\
Original task: [TASK]\\
Logged assistant response: [LOGGED RESPONSE]\\
Later user turn: [USER TURN]\\
\medskip
INSTRUCTIONS\\
-- Extract each explicit correction, criticism, clarification,
constraint, preference, desired answer property, or separate new
request.\\
-- Each component must express exactly one point and remain
understandable on its own.\\
-- Preserve the user's meaning and original order.\\
-- Do not infer requirements that are not stated.\\
-- Do not assign FIX/SPEC/NULL.\\
-- If no listed signal is present, return an empty list.\\
\medskip
Return JSON only:
\{"components": ["...", "..."]\}
\end{tcolorbox}
\caption{Prompt for atomic feedback-component extraction.}
\label{fig:prompt-extraction}
\end{figure}

\paragraph{Task-relative role assignment.}
The following prompt is applied independently to every extracted
component.

\begin{figure}[t]
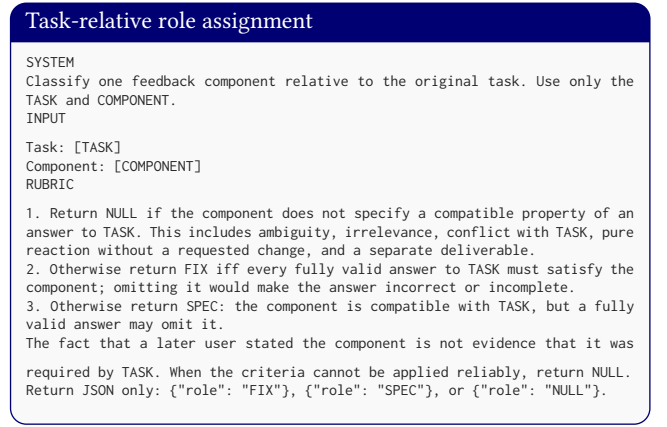

\centering
\begin{tcolorbox}[
  promptbox,
  title=Task-relative role assignment
]
\scriptsize\ttfamily
\color{black!85}
SYSTEM\\
Classify one feedback component relative to the original task. Use
only the TASK and COMPONENT.\\
\medskip
INPUT\\
Task: [TASK]\\
Component: [COMPONENT]\\
\medskip
RUBRIC\\
1. Return NULL if the component does not specify a compatible
property of an answer to TASK. This includes ambiguity, irrelevance,
conflict with TASK, pure reaction without a requested change, and a
separate deliverable.\\
2. Otherwise return FIX iff every fully valid answer to TASK must
satisfy the component; omitting it would make the answer incorrect or
incomplete.\\
3. Otherwise return SPEC: the component is compatible with TASK, but
a fully valid answer may omit it.\\
\medskip
The fact that a later user stated the component is not evidence that
it was required by TASK. When the criteria cannot be applied
reliably, return NULL.\\
\medskip
Return JSON only:
\{"role": "FIX"\}, \{"role": "SPEC"\}, or \{"role": "NULL"\}.
\end{tcolorbox}
\caption{Prompt for task-relative role assignment.}
\label{fig:prompt-role-assignment}
\end{figure}

\begin{figure}[t]
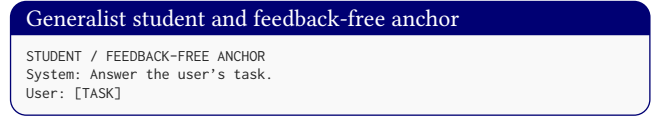

\centering
\begin{tcolorbox}[
  promptbox,
  title=Generalist student and feedback-free anchor
]
\scriptsize\ttfamily
\color{black!85}
STUDENT / FEEDBACK-FREE ANCHOR\\
System: Answer the user's task.\\
User: [TASK]
\end{tcolorbox}
\caption{Task-only serialization used by the Generalist student and
frozen feedback-free anchor in separate forward passes.}
\label{fig:prompt-generalist-student-anchor}
\end{figure}

\begin{figure}[t]
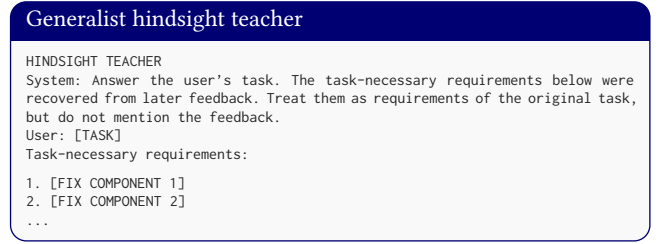

\centering
\begin{tcolorbox}[
  promptbox,
  title=Generalist hindsight teacher
]
\scriptsize\ttfamily
\color{black!85}
HINDSIGHT TEACHER\\
System: Answer the user's task. The task-necessary requirements below
were recovered from later feedback. Treat them as requirements of the
original task, but do not mention the feedback.\\
User: [TASK]\\
\medskip
Task-necessary requirements:\\
1. [FIX COMPONENT 1]\\
2. [FIX COMPONENT 2]\\
...
\end{tcolorbox}
\caption{Feedback-conditioned serialization used by the frozen
hindsight teacher.}
\label{fig:prompt-generalist-teacher}
\end{figure}

\paragraph{Generalist serializations.}
For $C_i^F=(c_{i1},\ldots,c_{im})$, we preserve component order and
number the components without paraphrasing them. The Generalist student
and feedback-free anchor use the task-only serialization in
Figure~\ref{fig:prompt-generalist-student-anchor} in separate forward
passes. The hindsight teacher uses
Figure~\ref{fig:prompt-generalist-teacher}. In all three passes, the
same logged prefix $y_{i,<t}^{\mathrm{log}}$ is appended by teacher
forcing.

\paragraph{Offline Specialist-target construction.}
This is the only Specialist-related prompt that receives $C_i^S$.
The \textsc{Spec} components are privileged offline information and
are not included in the Specialist's training or test-time input.

\begin{figure}[t]
\centering
\begin{tcolorbox}[
  promptbox,
  title=Offline Specialist-target construction
]
\scriptsize\ttfamily
\color{black!85}
SYSTEM\\
Construct one training target for a Specialist. SPEC components are
privileged offline information used to define candidate refinements;
they are unavailable to the Specialist.\\
\medskip
INPUT\\
Task: [TASK]\\
Generalist response: [GENERALIST RESPONSE]\\
SPEC components, in original order: [SPEC COMPONENTS]\\
\medskip
INSTRUCTIONS\\
-- For each SPEC component, separate the candidate refinement from
evidence that it applies: the component defines the candidate, but only
TASK and GENERALIST RESPONSE may justify its applicability.\\
-- Retain a candidate only when observable cues in TASK or GENERALIST
RESPONSE make it relevant without assuming an unobserved user
preference. Do not use the fact that it appeared in feedback as such a
cue.\\
-- Do not retain concrete conditions or details whose relevance is
supported only by the SPEC component.\\
-- Among the retained candidates, keep only those that are compatible
with all explicit task constraints, mutually compatible, and not
already realized in meaning by the Generalist response.\\
-- If no candidate remains both observably applicable and unmet, return
KEEP with no guidance.\\
-- Otherwise return APPLY and state only the minimal residual changes
needed for the retained, unmet refinements, preserving their original
order.\\
-- Do not write a replacement response.\\
\medskip
Return JSON only, using one of these schemas:\\
\{"action": "KEEP", "guidance": []\}\\
\{"action": "APPLY", "guidance": ["...", "..."]\}
\end{tcolorbox}
\caption{Prompt for constructing offline Specialist targets.}
\label{fig:prompt-specialist-target}
\end{figure}


\paragraph{Specialist input and target serialization.}
After offline target construction, the Specialist receives only the
task and Generalist response. The action is constrained to
$\{\texttt{KEEP},\texttt{APPLY}\}$, and residual guidance is generated
only after \texttt{APPLY}.

\begin{figure}[t]
\centering
\begin{tcolorbox}[
  promptbox,
  title=Specialist input and target serialization
]
\scriptsize\ttfamily
\color{black!85}
SYSTEM\\
Given a task and a Generalist response, decide whether an observably
applicable learned refinement remains unmet. Use only the task and
response, and do not assume an unobserved user preference. Do not
rewrite the response.\\
\medskip
INPUT\\
Task: [TASK]\\
Generalist response: [GENERALIST RESPONSE]\\
\medskip
OUTPUT\\
Return JSON only:\\
\{"action": "KEEP", "guidance": []\}\\
or\\
\{"action": "APPLY",
"guidance": ["minimal residual change", "..."]\}
\end{tcolorbox}
\caption{Prompt observed by the Specialist during training and
inference.}
\label{fig:prompt-specialist}
\end{figure}

\paragraph{Residual integration.}
This prompt is used only after an \texttt{APPLY} decision. The frozen
Generalist performs one integration pass.

\begin{figure}[t]
\centering
\begin{tcolorbox}[
  promptbox,
  title=Residual integration
]
\scriptsize\ttfamily
\color{black!85}
SYSTEM\\
Revise the Generalist response once so that it implements the residual
guidance.\\
\medskip
INPUT\\
Task: [TASK]\\
Generalist response: [GENERALIST RESPONSE]\\
Residual guidance: [GUIDANCE]\\
\medskip
INSTRUCTIONS\\
-- Implement only the changes in GUIDANCE.\\
-- Preserve all unaffected content, claims, structure, formatting, and
level of detail.\\
-- Do not add requirements not supported by TASK or GUIDANCE.\\
-- Return the complete final response only.
\end{tcolorbox}
\caption{Prompt for the frozen Generalist's single residual-integration
pass.}
\label{fig:prompt-integration}
\end{figure}

\subsection{Optimization}
\label{app:optimization-details}

Each training run uses four GPUs and bfloat16 precision. For trainable
methods, results reported over five runs use five fixed training seeds;
within a run, all compared trainable variants use the same data
manifest, training seed, and generation seed. Base has no training
seed; its five runs independently restart the complete evaluation under
identical greedy decoding and fixed seeds, so the resulting variation
reflects multi-GPU inference/runtime nondeterminism rather than
sampling. We use
AdamW and apply LoRA to the attention $q/k/v/o$ projections and the
MLP gate/up/down projections. Table~\ref{tab:slift-training-config}
reports the settings shared across backbones and training sources.

\begin{table}[!t]
\centering
\caption{Shared training configuration. ``Mean'' denotes length
normalization in the corresponding objective.}
\label{tab:slift-training-config}
\footnotesize
\setlength{\tabcolsep}{3pt}
\renewcommand{\arraystretch}{1.08}
\begin{tabularx}{\columnwidth}{@{}lXX@{}}
\toprule
\textbf{Setting} & \textbf{Generalist} & \textbf{Specialist} \\
\midrule
Trainable parameters
& $A_G$ only
& $A_S$ only \\
LoRA rank / scale
& $64/128$
& $16/32$ \\
LoRA dropout
& $0.05$
& $0.05$ \\
Learning rate
& $2\times10^{-5}$
& $2\times10^{-5}$ \\
Training epochs
& $2$
& $2$ \\
Behavioral anchor
& $\lambda_B=0.5$
& -- \\
Effective batch size
& $64$
& $32$ \\
Sampling
& all $C_i^F\neq\varnothing$
& $1{:}1$ \texttt{KEEP}/\texttt{APPLY} \\
Loss normalization
& response mean
& completion-token mean \\
Teacher supervision
& detached distributions
& offline targets \\
\bottomrule
\end{tabularx}
\end{table}

The Generalist uses logged prefixes only; it does not generate a
revised target response or use preference pairs or rewards. Both KL
terms are computed from complete next-token distributions and averaged
over response positions. The Specialist loss is computed over all
completion tokens in the balanced structured targets and normalized
by their total count. The Specialist adapter is
initialized independently from the frozen backbone, not from the
trained Generalist adapter.

Source-specific context limits and the remaining optimizer fields,
including the scheduler, warmup, weight decay, optimizer-step count,
and token budget, are recorded in the released run configurations. No
benchmark score is used in either training objective.

\subsection{Generation and Inference}

For a fixed backbone and training source, all compared methods use the
same chat template, context budget, output budget, and decoding
configuration. Qwen3-8B remains in non-thinking mode throughout.
WildFB-trained models use greedy decoding\newline
($\mathrm{temperature}=0$, $\mathrm{top\_p}=1$) for Base, Generalist,
Specialist, and residual-integration calls. MemoryBench uses the
official generation configuration and partition-specific output limits.
Specialist action decoding is constrained to
$\{\texttt{KEEP},\texttt{APPLY}\}$. A \texttt{KEEP} result returns
$y^G$ unchanged; an \texttt{APPLY} result triggers one frozen
Generalist integration call and no iterative editing.

\subsection{Cost Measurement}
\label{app:cost-measurement}

The training-cost panel in Figure~\ref{fig:efficiency} reports
GPU-hours for training-signal construction and parameter optimization.
For retrieval and external-memory systems, the panel marks that no
parameter-update stage is performed; it should not be interpreted as
zero cost for memory writing or index construction. Test-time cost is
measured end to end per test task and normalized by Base. It includes
all retrieval, memory access, module selection, Specialist, and model
generation calls required by each method. All methods are measured
with the same Qwen3-8B deployment and decoding budgets.

\subsection{Official Evaluation Implementations}

We use the official evaluation and scoring implementations without
modifying their benchmark definitions. The corresponding public
resources are listed in
Table~\ref{tab:official-evaluation-implementations}.

\begin{table}[!t]
\centering
\caption{Official implementations used for evaluation.}
\label{tab:official-evaluation-implementations}
\footnotesize
\setlength{\tabcolsep}{4pt}
\renewcommand{\arraystretch}{1.12}
\begin{tabularx}{\columnwidth}{
    @{}
    >{\raggedright\arraybackslash}p{0.25\columnwidth}
    >{\raggedright\arraybackslash}X
    @{}
}
\toprule
\textbf{Benchmark} & \textbf{Official resource} \\
\midrule
MemoryBench
& \url{https://github.com/THUIR/MemoryBench} \\
WildFB and WildReward-8B
& \url{https://github.com/THU-KEG/WildReward} \\
IFEval
& \url{https://github.com/google-research/google-research/tree/master/instruction_following_eval} \\
AlpacaEval 2.0
& \url{https://github.com/tatsu-lab/alpaca_eval} \\
MMLU-Pro
& \url{https://github.com/TIGER-AI-Lab/MMLU-Pro} \\
\bottomrule
\end{tabularx}
\end{table}

\subsection{Baseline Implementations}
\label{app:baseline-implementation}

\paragraph{Common protocol.}
For each backbone and dataset, all baselines receive exactly the same
training split. We represent each training interaction as
$S_i=(x_i,y_i^{\mathrm{log}},u_i)$, where $x_i$ is the original task and
its available task context, $y_i^{\mathrm{log}}$ is the logged model
response, and $u_i$ is the subsequent user turn. Retrieval indices,
memory states, preference data, and parameter updates are constructed
only from the training split. They are fixed before evaluation, and
responses generated for one test request are not written back to the
state used by later requests.

All methods use the same backbone, system prompt, chat template,
context-length budget, and maximum output length. Qwen3-8B is evaluated
in non-thinking mode. Following the benchmark-specific evaluation
protocol, MemoryBench uses the official generation configuration,
including temperature $0.1$ and its partition-specific output limits,
whereas WildFB and the general-purpose evaluation sets use greedy
decoding. Base directly applies the corresponding unadapted backbone and
receives no historical interaction information.

\paragraph{Direct retrieval.}
We treat each complete training
interaction as one retrievable entry. The original task $x_i$, rather
than the feedback turn alone, is used as the retrieval key, while the
corresponding task, logged response, and subsequent user turn are
jointly retained as the retrieved content. Given a test request,
BM25~\cite{robertson2009probabilistic} ranks the training interactions
by sparse lexical relevance. The dense variant encodes the training
tasks and the test request using Qwen3-Embedding-0.6B
~\cite{zhang2025qwen3embedding} and ranks sessions by embedding
similarity. Neither variant uses an additional reranker.

For both retrievers, the top-$5$ training interactions are serialized
in descending retrieval order and prepended to the current request as
historical examples. If the resulting input exceeds the model context
window, we reserve the current request and generation budget and use
a bisection-based truncation procedure to reduce only the
retrieved portion. Thus, both retrieval methods use the same number
and format of candidate sessions and differ only in their retrieval
function.

\paragraph{Memory systems.}
All memory-system baselines receive the same training-interaction stream
in its original order. Each training interaction is inserted once,
after which the resulting memory state is frozen for evaluation. We
retain each method's official extraction, consolidation, linking, and
update operations rather than replacing them with a shared memory
representation. Whenever a method exposes a retrieval-depth parameter,
we use the top-$5$ memory entries, matching the retrieval budget above
across all methods.

For A-Mem~\cite{xu2025amemagenticmemoryllm}, each training interaction
is converted into a structured note and processed by its official
note-linking and memory-evolution operations. For
MemoryOS~\cite{kang-etal-2025-memory}, training interactions are
sequentially written into its short-, mid-, and long-term hierarchy,
with the official promotion and consolidation procedure applied during
memory construction. We follow the corresponding MemoryBench
configurations for these two systems.

For MemOS~\cite{li2025memos}, we use its official memory-writing and
retrieval interfaces but replace its default response generator, when
necessary, with the same backbone used by all other methods. For
ReMem~\cite{wei2025evomemory}, we apply its released memory refinement
procedure to the training interactions and retain the evolved memory
for subsequent test requests. In every case, retrieved memories are
passed to the common backbone through the method's official prompt
template, and the final response is generated under the shared
decoding configuration.

\paragraph{Preference-data construction.}
SFT and DPO require target responses or preference pairs that are not
present in raw interaction logs. We therefore apply the following
preprocessing protocol. For each training
interaction, the corresponding frozen backbone distills actionable
information from $(x_i,y_i^{\mathrm{log}},u_i)$ into a semi-structured
rule set $R_i$; interactions with no non-empty rule set are discarded.
The same backbone then revises the logged response conditioned on
$u_i$ and $R_i$ to produce $y_i^+$, while the logged response is
retained as $y_i^-=y_i^{\mathrm{log}}$. This yields
\begin{equation}
\mathcal D_{\mathrm{pref}}
=
\left\{(x_i,y_i^+,y_i^-)\right\}.
\end{equation}
The same pairs are used by SFT and DPO. SFT and DPO do not
receive the feedback, rule set, or
\textsc{Fix}/\textsc{Spec}/\textsc{Null} labels as model inputs.

\paragraph{SFT and DPO}
SFT~\cite{ouyang2022training} trains one global LoRA adapter on
$\{(x_i,y_i^{+})\}$ using completion-only negative log-likelihood.
DPO~\cite{rafailov2024directpreferenceoptimizationlanguage} trains one
global adapter on the complete chosen--rejected pairs, with the
unadapted backbone serving as the frozen reference policy and
$\beta=0.1$. Each trains a single global adapter and retains no
external state at inference time.

Both methods use LoRA rank $64$
and dropout $0.05$. We train for eight epochs with a learning rate of
$5\times10^{-4}$ and a cosine schedule. Training uses
\texttt{bfloat16}, per-device batch size $1$ on four GPUs, and gradient
accumulation over eight steps, giving an effective batch size of $32$.
We use Adam coefficients
$(\beta_1,\beta_2)=(0.9,0.98)$, $\epsilon=10^{-8}$, weight decay
$10^{-4}$, no learning-rate warm-up, and gradient-norm clipping at
$1.0$. The maximum sequence length is replaced with the same
backbone- and benchmark-specific context budget used by the other
methods. For each run, we select the epoch with the lowest validation
cross-entropy on the chosen responses. At inference time, the selected
adapter generates directly from the task input without access to the
training logs or constructed preference data.

\paragraph{SDPO}
For SDPO~\cite{kleinebuening2026aligning}, we use its offline,
off-policy formulation for logged user interactions. Each training
interaction provides a tuple $(x_i,y_i^{\mathrm{log}},u_i)$. At every
logged response position $t$, the student distribution observes only
the original task and logged prefix, whereas the stop-gradient
hindsight distribution additionally observes the subsequent user turn:
\begin{align}
q_{i,t}
&=
\pi_{\theta}
\left(
\cdot
\mid
x_i,
y_{i,<t}^{\mathrm{log}}
\right), \\
p_{i,t}^{\mathrm{fb}}
&=
\operatorname{sg}\!\left[
\pi_{\theta}
\left(
\cdot
\mid
x_i,
u_i,
y_{i,<t}^{\mathrm{log}}
\right)
\right].
\end{align}
The adapter minimizes
\begin{equation}
\mathcal L_{\mathrm{SDPO}}
=
\frac{1}{|\mathcal D|}
\sum_i
\frac{1}{|y_i^{\mathrm{log}}|}
\sum_{t=1}^{|y_i^{\mathrm{log}}|}
D_{\mathrm{KL}}
\left(
q_{i,t}
\,\middle\|\,
p_{i,t}^{\mathrm{fb}}
\right).
\label{eq:baseline-sdpo}
\end{equation}
The logged response supplies all scored prefixes, so SDPO does not
sample new responses, construct chosen--rejected pairs, or invoke a
reward model.

For parameter-count comparability, we apply this objective through one
LoRA adapter with rank $64$ and dropout $0.05$. Following the original
logged-interaction experiments, we use a learning rate of
$2\times10^{-6}$, an effective batch size of $32$, two epochs, a
$5\%$ warm-up ratio, a cosine learning-rate schedule, and an
8-bit AdamW optimizer. We replace the original fixed prompt and
completion limits with our shared context and output budgets so that
long-context MemoryBench instances are evaluated without an additional
baseline-specific truncation.

\section{Additional Experiments}
\label{app:additional-experiments}

\subsection{Analysis Settings}
\label{app:analysis-settings}

\paragraph{Judge protocol.}
The task/feedback analysis and the audits below use a fixed,
source-blind \texttt{qwen3.7-max} snapshot~\cite{qwen37}, independent
of the two evaluated backbones. We use deterministic decoding,
request JSON only, and hide the training source, backbone, benchmark
name, dataset labels, downstream model output, and evaluation result.
Malformed outputs are retried with the same prompt; records with
incomplete provider output after retry are excluded.

\paragraph{Task and feedback analysis.}
We select the first feedback opportunity for each root training task.
MemoryBench provides backbone-specific response and feedback logs; a
fixed seed assigns each root task to one of the two backbone logs so
that each task contributes one observation. WildFB contributes one
shared real-user log per root task. Three annotation calls preserve the
information boundary: the task call receives the task context $H$; the
response call receives $(H,y_i^{\mathrm{log}})$; and the feedback call
receives $(H,y_i^{\mathrm{log}},u_i)$. The four reported feedback
criteria are response relevance; relevance plus diagnostic support;
relevance plus specificity; and full usability, which additionally
requires actionability.

For standardized comparisons, we stratify root tasks by task operation
and objective verifiability, retain strata with at least 20 tasks from
each source, and reweight both sources to the equal-weight average of
their distributions over shared strata. We use 2,000 root-task
bootstrap replicates and report percentile 95\% confidence intervals.

\subsection{Complete Evaluation Protocol}
\label{app:complete-evaluation-protocol}

\paragraph{MemoryBench.}
We evaluate 150/143/150/275 held-out requests in the Short--Short
(SS), Short--Long (SL), Long--Short (LS), and Long--Long (LL)
partitions, respectively. We retain each constituent dataset's native
scorer and the fixed normalization statistics distributed with the
official harness~\cite{ai2026memorybenchbenchmarkmemorycontinual}. Raw scores are normalized
within each constituent dataset and then aggregated over test
instances. We report the resulting sample-weighted min--max score
(Norm-Score, multiplied by 100) and Z-score. The Average column is the
unweighted mean of the four partition-level Norm-Scores. The official
generation limits are 2,048 new tokens for SS, SL, and LS and 12,000
for LL. All compared methods use identical test keys, prompts, output
limits, and evaluator configurations.

\paragraph{WildFB and capability evaluations.}
All formal generations use the common greedy configuration specified
in \S~\ref{app:impl}. IFEval is evaluated on all 541 official prompts;
a prompt receives prompt-level strict credit only when every
verifiable instruction attached to it passes the corresponding
checker~\cite{zhou2023instruction}. AlpacaEval~2.0 uses all 805
official instructions, the official GPT-4-Turbo reference responses
and weighted auto-annotator, and reports the length-controlled (LC)
win rate~\cite{dubois2025lengthcontrolledalpacaevalsimpleway}. MMLU-Pro uses all 12,032 test
questions with the official five-shot chain-of-thought protocol and
reports exact accuracy after extracting the final A--J
choice~\cite{wang2024mmlupro}.

For the in-domain comparison, 4,929 held-out WildFB interactions remain
after common structural validity checks. WildReward-8B scores the
generated and logged responses under the same task context and
query. For ordinal logits $\ell_{ik}$, its scalar score is
\begin{equation}
r_i = 1+\sum_k \sigma(\ell_{ik}).
\end{equation}
We report
$100N^{-1}\sum_i\mathbf{1}[r_i^{\mathrm{gen}}>
r_i^{\mathrm{log}}]$; exact ties are non-wins. This is a paired
cross-policy comparison against the logged response, rather than an
absolute measure of response quality. No WildFB satisfaction label is
used in generation, scoring, target construction, or threshold
selection.

\begin{table*}[!t]
\centering
\caption{Expanded ablation results across two backbones. The first
five columns report MemoryBench Norm-Score, while the last four report
IFEval Prompt-Strict, AlpacaEval~2.0 LC win rate, MMLU-Pro accuracy,
and WildReward win rate after training on WildFB logs. All entries are
means over five runs, with standard deviations in parentheses.}
\label{tab:full-ablation-results}

\footnotesize
\setlength{\tabcolsep}{1.7pt}
\renewcommand{\arraystretch}{1.10}

\begin{tabularx}{\textwidth}{
    @{}
    >{\raggedright\arraybackslash}X
    *{4}{>{\centering\arraybackslash}p{0.060\textwidth}}
    >{\centering\arraybackslash}p{0.065\textwidth}
    >{\centering\arraybackslash}p{0.075\textwidth}
    >{\centering\arraybackslash}p{0.075\textwidth}
    >{\centering\arraybackslash}p{0.085\textwidth}
    >{\centering\arraybackslash}p{0.090\textwidth}
    @{}
}
\toprule
&
\multicolumn{5}{c}{\textbf{MemoryBench-trained}}
&
\multicolumn{4}{c}{\textbf{WildFB-trained}} \\
\cmidrule(lr){2-6}
\cmidrule(lr){7-10}

\textbf{Variant}
& \textbf{SS}
& \textbf{SL}
& \textbf{LS}
& \textbf{LL}
& \textbf{Avg.}
& \textbf{IFEval}
& \textbf{Alpaca}
& \textbf{MMLU-Pro}
& \textbf{WildReward} \\
\midrule

\multicolumn{10}{@{}l}{\textbf{Qwen3-8B}} \\[-1pt]

\textbf{\method{}}
& \eststd{\textbf{70.49}}{.43}
& \eststd{\textbf{58.00}}{.37}
& \eststd{\textbf{45.22}}{.40}
& \eststd{\textbf{49.68}}{.45}
& \eststd{\textbf{55.85}}{.25}
& \eststd{\textbf{86.44}}{.94}
& \eststd{\textbf{47.68}}{.24}
& \eststd{\textbf{62.15}}{.10}
& \eststd{\textbf{78.76}}{.17} \\

\hspace{0.6em}w/o Atomic Extraction
& \eststd{67.31}{.52} & \eststd{56.84}{.48}
& \eststd{44.07}{.47} & \eststd{43.58}{.55}
& \eststd{52.95}{.31} & \eststd{83.78}{.86}
& \eststd{45.51}{.23} & \eststd{62.04}{.09}
& \eststd{78.08}{.21} \\

\hspace{0.6em}w/o Task Context for Roles
& \eststd{65.42}{.58} & \eststd{53.87}{.54}
& \eststd{43.06}{.56} & \eststd{41.93}{.63}
& \eststd{51.07}{.35} & \eststd{83.01}{.94}
& \eststd{44.31}{.25} & \eststd{61.92}{.11}
& \eststd{77.71}{.24} \\

\hspace{0.6em}w/o Separate Pathways
& \eststd{66.72}{.55} & \eststd{55.46}{.46}
& \eststd{45.08}{.53} & \eststd{41.86}{.60}
& \eststd{52.28}{.33} & \eststd{82.94}{.89}
& \eststd{44.63}{.22} & \eststd{62.19}{.08}
& \eststd{77.84}{.22} \\

\hspace{0.6em}w/o Specialist (\method{}-Gen)
& \eststd{65.25}{.49} & \eststd{55.33}{.39}
& \eststd{43.20}{.49} & \eststd{43.72}{.48}
& \eststd{51.88}{.28} & \eststd{84.40}{.83}
& \eststd{47.02}{.22} & \eststd{62.15}{.11}
& \eststd{77.46}{.19} \\

\hspace{0.6em}w/o \texttt{KEEP} Supervision
& \eststd{64.49}{.66} & \eststd{54.43}{.58}
& \eststd{42.71}{.69} & \eststd{43.33}{.72}
& \eststd{51.24}{.43} & \eststd{72.18}{1.28}
& \eststd{46.38}{.31} & \eststd{59.47}{.18}
& \eststd{74.63}{.30} \\

\midrule

\multicolumn{10}{@{}l}{\textbf{Ministral3-14B}} \\[-1pt]

\textbf{\method{}}
& \eststd{\textbf{67.81}}{.45}
& \eststd{\textbf{39.22}}{.58}
& \eststd{\textbf{48.75}}{.53}
& \eststd{\textbf{57.09}}{.47}
& \eststd{\textbf{53.22}}{.31}
& \eststd{\textbf{69.71}}{1.02}
& \eststd{\textbf{66.35}}{.21}
& \eststd{\textbf{52.20}}{.08}
& \eststd{\textbf{81.10}}{.14} \\

\hspace{0.6em}w/o Atomic Extraction
& \eststd{65.74}{.55} & \eststd{39.08}{.66}
& \eststd{45.29}{.59} & \eststd{55.09}{.61}
& \eststd{51.30}{.34} & \eststd{67.89}{.91}
& \eststd{65.22}{.19} & \eststd{52.08}{.10}
& \eststd{80.61}{.18} \\

\hspace{0.6em}w/o Task Context for Roles
& \eststd{63.88}{.61} & \eststd{38.94}{.70}
& \eststd{47.37}{.57} & \eststd{52.77}{.67}
& \eststd{50.74}{.36} & \eststd{67.34}{.74}
& \eststd{63.98}{.22} & \eststd{51.94}{.12}
& \eststd{80.54}{.21} \\

\hspace{0.6em}w/o Separate Pathways
& \eststd{64.57}{.58} & \eststd{39.11}{.62}
& \eststd{46.32}{.60} & \eststd{54.20}{.64}
& \eststd{51.05}{.33} & \eststd{67.62}{.86}
& \eststd{64.61}{.18} & \eststd{52.27}{.09}
& \eststd{80.43}{.20} \\

\hspace{0.6em}w/o Specialist (\method{}-Gen)
& \eststd{64.23}{.51} & \eststd{39.20}{.59}
& \eststd{46.61}{.47} & \eststd{55.62}{.54}
& \eststd{51.42}{.32} & \eststd{67.28}{.79}
& \eststd{64.27}{.17} & \eststd{52.20}{.09}
& \eststd{80.79}{.16} \\

\hspace{0.6em}w/o \texttt{KEEP} Supervision
& \eststd{63.54}{.67} & \eststd{38.68}{.73}
& \eststd{45.79}{.68} & \eststd{55.43}{.64}
& \eststd{50.86}{.44} & \eststd{57.63}{1.36}
& \eststd{64.06}{.28} & \eststd{49.38}{.19}
& \eststd{75.81}{.29} \\

\bottomrule
\end{tabularx}
\end{table*}

\subsection{Complete WildFB Results}
\label{app:wildfb-results-full}

\begin{table}[!t]
\centering
\caption{Complete results for models trained on WildFB, underlying
Figure~\ref{fig:wildfb}. All methods use the common generation and
evaluation protocol in
\S~\ref{app:complete-evaluation-protocol}. Results are means over five
runs, with standard deviations in parentheses. Base has no training
seed; its five runs independently restart the complete fixed-seed
greedy evaluation.}
\label{tab:wildfb-results-full}
\footnotesize
\setlength{\tabcolsep}{2.3pt}
\renewcommand{\arraystretch}{1.06}
\begin{tabular*}{\columnwidth}{
@{\extracolsep{\fill}}lrrrr@{}}
\toprule
\textbf{Method}
& \shortstack{\textbf{IFEval}\\\textit{Prompt-Strict}}
& \shortstack{\textbf{AlpacaEval}\\\textbf{2.0}\\\textit{LC win rate}}
& \shortstack{\textbf{MMLU-Pro}\\\textit{Accuracy}}
& \shortstack{\textbf{WildReward}\\\textit{Win rate}} \\
\midrule

\multicolumn{5}{@{}l}{\textbf{Qwen3-8B}} \\
Base
& \eststd{82.55}{.17} & \eststd{44.51}{.26}
& \eststd{62.32}{.12} & \eststd{77.36}{.18} \\
SFT
& \eststd{81.81}{.74} & \eststd{43.37}{.28}
& \eststd{61.68}{.14} & \eststd{76.88}{.23} \\
DPO
& \eststd{82.03}{.69} & \eststd{43.84}{.25}
& \eststd{61.92}{.16} & \eststd{77.13}{.21} \\
SDPO
& \eststd{79.82}{.91} & \eststd{41.76}{.31}
& \eststd{60.47}{.19} & \eststd{75.68}{.27} \\
\method{}-Gen
& \eststd{84.40}{.83} & \eststd{47.02}{.22}
& \eststd{62.15}{.11} & \eststd{77.46}{.19} \\
\textbf{\method{}}
& \eststd{\textbf{86.44}}{.94}
& \eststd{\textbf{47.68}}{.24}
& \eststd{62.15}{.10}
& \eststd{\textbf{78.76}}{.17} \\

\midrule
\multicolumn{5}{@{}l}{\textbf{Ministral3-14B}} \\
Base
& \eststd{65.92}{.43} & \eststd{63.83}{.20}
& \eststd{52.31}{.10} & \eststd{80.02}{.16} \\
SFT
& \eststd{64.73}{.68} & \eststd{62.47}{.25}
& \eststd{51.63}{.13} & \eststd{79.68}{.21} \\
DPO
& \eststd{65.03}{.72} & \eststd{62.93}{.23}
& \eststd{51.84}{.14} & \eststd{79.83}{.19} \\
SDPO
& \eststd{63.51}{.88} & \eststd{60.76}{.29}
& \eststd{50.47}{.17} & \eststd{78.17}{.26} \\
\method{}-Gen
& \eststd{67.28}{.79} & \eststd{64.27}{.17}
& \eststd{52.20}{.09} & \eststd{80.79}{.16} \\
\textbf{\method{}}
& \eststd{\textbf{69.71}}{1.02}
& \eststd{\textbf{66.35}}{.21}
& \eststd{52.20}{.08}
& \eststd{\textbf{81.10}}{.14} \\

\bottomrule
\end{tabular*}
\end{table}

Across both backbones, \method{} obtains the highest IFEval,
AlpacaEval~2.0, and WildReward scores among the compared methods.
The largest gains occur on instruction following and open-ended
response quality, while MMLU-Pro remains within $0.17$ points of Base
for Qwen3-8B and $0.11$ points for Ministral3-14B.

\subsection{Full Ablation Results}
\label{app:full-ablation-results}

The expanded results use the same five variants as
Table~\ref{tab:ablation-slift}. \emph{w/o Atomic Extraction} assigns
one role to the complete feedback turn. \emph{w/o Task Context for
Roles} removes $x_i$ from role assignment. \emph{w/o Separate
Pathways} sends both \textsc{Fix} and \textsc{Spec} through the
Generalist while continuing to exclude \textsc{Null}. \emph{w/o
Specialist} discards \textsc{Spec} supervision, and \emph{w/o
\texttt{KEEP} Supervision} trains the Specialist only on
\texttt{APPLY} targets.

The two-backbone results match the main ablation trends. Removing the
task context from role assignment produces the largest MemoryBench
average and AlpacaEval~2.0 losses on both backbones. Sending
\textsc{Fix} and \textsc{Spec} through one Generalist or discarding
\textsc{Spec} both underperform the complete model. Removing
\texttt{KEEP} supervision produces the largest IFEval loss on both
backbones.

\subsection{Full Online Evolution Results}
\label{app:online-evolution-full}

We split the MemoryBench training tasks into two equal, disjoint
batches and keep the test set fixed. Iteration~0 is the shared
Qwen3-8B Base. At iteration~1, each method is trained on the
pre-collected logs from batch~1. The resulting model is then deployed
on batch~2, where the MemoryBench User Simulator generates
method-specific on-policy feedback. Iteration~2 applies the
corresponding incremental update.

\begin{table}[!t]
\centering
\caption{Online evolution on MemoryBench with Qwen3-8B. Each entry is
the macro-average Norm-Score over SS/SL/LS/LL. Iteration columns report
five-run means with standard deviations in parentheses; delta columns
report paired changes between adjacent iterations.}
\label{tab:online-evolution-full}
\footnotesize
\setlength{\tabcolsep}{1.4pt}
\renewcommand{\arraystretch}{1.06}
\begin{tabular*}{\columnwidth}{
@{\extracolsep{\fill}}lrrrrr@{}}
\toprule
\textbf{Method}
& \textbf{Iter.~0} & \textbf{Iter.~1} & \textbf{Iter.~2}
& \textbf{$\Delta_{0\to1}$} & \textbf{$\Delta_{1\to2}$} \\
\midrule
SFT
& \eststd{49.17}{.28} & \eststd{49.75}{.35}
& \eststd{50.08}{.37}
& \eststd{+0.58}{.22} & \eststd{+0.33}{.19} \\
SDPO
& \eststd{49.17}{.28} & \eststd{50.05}{.38}
& \eststd{50.55}{.40}
& \eststd{+0.88}{.25} & \eststd{+0.50}{.21} \\
\method{}-Gen
& \eststd{49.17}{.28} & \eststd{52.72}{.30}
& \eststd{54.34}{.32}
& \eststd{+3.55}{.17} & \eststd{+1.62}{.19} \\
\textbf{\method{}}
& \eststd{49.17}{.28} & \eststd{\textbf{53.08}}{.27}
& \eststd{\textbf{54.86}}{.30}
& \eststd{\textbf{+3.91}}{.16}
& \eststd{\textbf{+1.78}}{.18} \\
\bottomrule
\end{tabular*}
\end{table}

All methods improve after both training rounds. \method{} obtains the
highest score at each trained iteration and the largest increment in
each round, while \method{}-Gen remains below the complete model.

\subsection{Feedback Processing and Specialist-Target Audit}
\label{app:scope-control}

\paragraph{Extraction and role assignment.}
The extractor observes $(x_i,y_i^{\mathrm{log}},u_i)$ and returns an
ordered list of minimal components, possibly empty. Role assignment is
a separate call that observes only $(x_i,c_{ij})$. It assigns
\textsc{Fix} when every fully valid response to $x_i$ must satisfy the
component, \textsc{Spec} when the component is task-compatible but
omissible, and \textsc{Null} when no reliable positive update is
supported.

The extraction audit samples 300 processed root tasks per source,
split evenly across the Qwen3-8B and Ministral3-14B processing
pipelines. The independent judge receives the logged interaction and
extracted list, but not component roles or downstream results, and
checks coverage, faithfulness, and atomicity. The role audit samples
500 components from each assigned role and source, split evenly across
the two processing backbones. The judge receives only $(x_i,c_{ij})$
and independently assigns a \textsc{Fix}-like, \textsc{Spec}-like,
\textsc{Null}-like, or uncertain label.

\begin{table}[!t]
\centering
\caption{Source-blind extraction audit. Values are percentages of
sampled root tasks ($n=300$ per source); ``All'' requires coverage,
faithfulness, and atomic/self-contained decomposition.}
\label{tab:extraction-audit}
\footnotesize
\setlength{\tabcolsep}{2.6pt}
\renewcommand{\arraystretch}{1.07}
\begin{tabular*}{\columnwidth}{
@{\extracolsep{\fill}}lrrrrr@{}}
\toprule
\textbf{Source} & \textbf{$n$} & \textbf{Coverage}
& \textbf{Faithful} & \textbf{Atomic} & \textbf{All} \\
\midrule
MemoryBench & 300 & 96.3 & 98.7 & 95.0 & 92.7 \\
WildFB      & 300 & 89.0 & 95.3 & 90.0 & 84.0 \\
\bottomrule
\end{tabular*}
\end{table}

\begin{table}[!t]
\centering
\caption{Source-blind role audit ($n=500$ per assigned-role/source
cell). Counts show the independent judge's task-relative label;
``Support'' is the diagonal percentage.}
\label{tab:scope-role-audit}
\footnotesize
\setlength{\tabcolsep}{1.6pt}
\renewcommand{\arraystretch}{1.08}
\begin{tabular*}{\columnwidth}{
@{\extracolsep{\fill}}llrrrrr@{}}
\toprule
\textbf{Source}
& \shortstack{\textbf{Assigned}\\\textbf{role}}
& \shortstack{\textbf{FIX-}\\\textbf{like}}
& \shortstack{\textbf{SPEC-}\\\textbf{like}}
& \shortstack{\textbf{NULL-}\\\textbf{like}}
& \shortstack{\textbf{Uncer-}\\\textbf{tain}}
& \shortstack{\textbf{Support}\\\textbf{(\%)}} \\
\midrule
\multirow{3}{*}{MemoryBench}
& \textsc{Fix}  & 461 & 23  & 10  & 6  & 92.2 \\
& \textsc{Spec} & 22  & 463 & 9   & 6  & 92.6 \\
& \textsc{Null} & 17  & 36  & 439 & 8  & 87.8 \\
\midrule
\multirow{3}{*}{WildFB}
& \textsc{Fix}  & 422 & 44  & 22  & 12 & 84.4 \\
& \textsc{Spec} & 45  & 416 & 23  & 16 & 83.2 \\
& \textsc{Null} & 29  & 33  & 426 & 12 & 85.2 \\
\bottomrule
\end{tabular*}
\end{table}

\begin{figure}[!t]
\centering
\begin{tcolorbox}[
  promptbox,
  colframe=teal!55!black,
  colback=teal!2!white,
  colbacktitle=teal!55!black,
  title=Case Study,
  width=\columnwidth
]
\scriptsize
\color{black!85}

\textbf{1. Benchmark input (verbatim)}\\
phrase an email to my manager kindly asking for feedback for a
dashboard I sent him for his use, and also ask him if I should publish
this dashboard to the team, and if there's a certain way he wants me
to publish it, or certain orientation to emphasis when publishing

\medskip
\textbf{2. Generalist response $y^G$}\\
\textbf{Subject: Feedback and Guidance on Dashboard Sharing}

Hi [Manager's Name],

I hope you're doing well. I'm following up on the dashboard I sent for
your use and would appreciate your feedback on its clarity, usefulness,
and any areas that could be improved. I'd also like to ask whether you
think I should publish it to the team. If so, please let me know whether
you have a preferred channel or format for sharing it and whether there
are particular findings or an overall orientation you would like me to
emphasize.

I'm happy to make any adjustments before sharing it. Thank you for your
guidance.

Best,\\
\textnormal{[Name]}

\medskip
\textbf{3. Specialist output}\\
{\ttfamily
\{"action": "APPLY",\\
\quad "guidance": ["Make the email easier for a busy manager to scan:
shorten the opening and present the publication-channel and emphasis
questions as two brief bullets. Preserve the professional tone and
every requested decision."]\}
}

\medskip
\textbf{4. Integrated response}\\
\textbf{Subject: Feedback and Guidance on Dashboard Sharing}

Hi [Manager's Name],

Could you please share your feedback on the dashboard I sent,
particularly anything that should be improved or clarified?

If you think it should be shared with the team, could you also advise:
\begin{itemize}
    \item which channel or format I should use; and
    \item which findings or overall orientation I should emphasize?
\end{itemize}

I'm happy to revise it before sharing. Thank you for your guidance.

Best,\\
\textnormal{[Name]}

\medskip
\textbf{5. Pairwise evaluation outcome}\\
The pairwise evaluator prefers the integrated response over the
Generalist response.

\end{tcolorbox}
\caption{An observed \texttt{APPLY} trace on AlpacaEval~2.0. The
Generalist response already satisfies every explicit request, while the
Specialist adds a compatible, nonessential refinement that improves
concision and presentation.}
\label{fig:specialist-case}
\vskip -0.1in
\end{figure}

\paragraph{Specialist targets.}
For each source, we sample 250 target-\texttt{KEEP} and 250
target-\texttt{APPLY} states, split evenly across the two backbone
pipelines. The judge receives $(x_i,y_i^G,C_i^S)$ but not the
constructed target. It independently determines whether the current
task--response state supports a missing compatible refinement and
therefore calls for \texttt{KEEP} or \texttt{APPLY}. For
\texttt{APPLY} states, a separate pass checks whether the guidance is
grounded in $C_i^S$, covers the missing compatible changes, and remains
residual rather than replacing the response. This audit evaluates the
offline targets rather than the trained Specialist itself.

\begin{table}[!t]
\centering
\caption{Source-blind Specialist-target audit. ``Target-consistent''
requires an independently supported action and, for \texttt{APPLY},
grounded, complete, residual-only guidance. Cells report counts with
percentages in parentheses.}
\label{tab:scope-state-audit}
\footnotesize
\setlength{\tabcolsep}{2.5pt}
\renewcommand{\arraystretch}{1.08}
\begin{tabularx}{\columnwidth}{
    @{}
    >{\raggedright\arraybackslash}X
    >{\centering\arraybackslash}p{0.21\columnwidth}
    >{\centering\arraybackslash}p{0.18\columnwidth}
    >{\centering\arraybackslash}p{0.22\columnwidth}
    @{}
}
\toprule
\textbf{Metric}
& \textbf{MemoryBench}
& \textbf{WildFB}
& \textbf{Pooled} \\
\midrule
\texttt{KEEP}/\texttt{APPLY}
& 250/250
& 250/250
& 500/500 \\

\texttt{KEEP} supported
& 236/250 (94.4)
& 221/250 (88.4)
& 457/500 (91.4) \\

\texttt{APPLY} supported
& 228/250 (91.2)
& 217/250 (86.8)
& 445/500 (89.0) \\

Action consistent
& 464/500 (92.8)
& 438/500 (87.6)
& 902/1,000 (90.2) \\

Guidance faithful
& 233/250 (93.2)
& 226/250 (90.4)
& 459/500 (91.8) \\

Target consistent
& 449/500 (89.8)
& 420/500 (84.0)
& 869/1,000 (86.9) \\
\bottomrule
\end{tabularx}
\end{table}

\begin{table}[!t]
\centering
\caption{Full stage endpoints on Ministral3-14B. Base, Generalist,
and Full are the scores used in Table~\ref{tab:stage-transfer}.
\texttt{APPLY} and Edited are recorded counts.}
\label{tab:stage-endpoints-full}
\footnotesize
\setlength{\tabcolsep}{1.7pt}
\renewcommand{\arraystretch}{1.07}
\begin{tabular*}{\columnwidth}{
@{\extracolsep{\fill}}lrrrrrr@{}}
\toprule
\textbf{Evaluation}
& \textbf{$N$}
& \textbf{Base}
& \shortstack{\textbf{Gener-}\\\textbf{alist}}
& \textbf{Full}
& \textbf{\texttt{APPLY}}
& \textbf{Edited} \\
\midrule

\multicolumn{7}{@{}l}{\textbf{MemoryBench training log}} \\
SS & 150 & 61.20 & 64.23 & 67.81 & 29 & 27 \\
SL & 143 & 37.86 & 39.20 & 39.22 &  2 &  2 \\
LS & 150 & 45.46 & 46.61 & 48.75 & 16 & 15 \\
LL & 275 & 54.71 & 55.62 & 57.09 & 25 & 23 \\

\midrule
\multicolumn{7}{@{}l}{\textbf{WildFB training log}} \\
IFEval
& 541 & 65.92 & 67.28 & 69.71 & 77 & 71 \\
AlpacaEval~2.0
& 805 & 63.83 & 64.27 & 66.35 & 103 & 95 \\
MMLU-Pro
& 12{,}032 & 52.31 & 52.20 & 52.20 & 0 & 0 \\
WildReward
& 4{,}929 & 80.02 & 80.79 & 81.10 & 123 & 108 \\

\bottomrule
\end{tabular*}
\end{table}

\begin{table}[!t]
\centering
\caption{\texttt{qwen3.7-max}-judged feedback yields. Values are
percentages of first feedback opportunities, with root-task bootstrap
95\% confidence intervals shown as subscripts. Standardized estimates
reweight both training logs to an equal-source pooled distribution over
task-operation and verifiability strata.}
\label{tab:task-feedback-usability-full}
\footnotesize
\setlength{\tabcolsep}{2.2pt}
\renewcommand{\arraystretch}{1.16}
\begin{tabularx}{\columnwidth}{
@{}>{\raggedright\arraybackslash}X
l
>{\centering\arraybackslash}p{0.19\columnwidth}
>{\centering\arraybackslash}p{0.19\columnwidth}
>{\centering\arraybackslash}p{0.20\columnwidth}@{}
}
\toprule
\textbf{Signal yield}
& \textbf{Log}
& \textbf{Observed}
& \textbf{Standardized}
& \textbf{WF$-$MB} \\
\midrule

\multirow{2}{=}{Response-relevant}
& MB
& \(98.1_{\scriptscriptstyle[97.1,\,99.0]}\)
& \(97.6_{\scriptscriptstyle[96.4,\,98.9]}\)
& \multirow{2}{*}{\(
-23.3_{\scriptscriptstyle[-25.6,\,-21.3]}
\)} \\
& WF
& \(73.2_{\scriptscriptstyle[72.5,\,73.9]}\)
& \(74.3_{\scriptscriptstyle[72.6,\,75.9]}\)
& \\

\addlinespace[2pt]

\multirow{2}{=}{Relevant and supported}
& MB
& \(97.4_{\scriptscriptstyle[96.3,\,98.4]}\)
& \(97.1_{\scriptscriptstyle[95.8,\,98.5]}\)
& \multirow{2}{*}{\(
-29.9_{\scriptscriptstyle[-32.4,\,-27.7]}
\)} \\
& WF
& \(64.0_{\scriptscriptstyle[63.2,\,64.8]}\)
& \(67.1_{\scriptscriptstyle[65.3,\,68.9]}\)
& \\

\addlinespace[2pt]

\multirow{2}{=}{Relevant and specific}
& MB
& \(97.7_{\scriptscriptstyle[96.6,\,98.7]}\)
& \(97.3_{\scriptscriptstyle[96.0,\,98.6]}\)
& \multirow{2}{*}{\(
-39.5_{\scriptscriptstyle[-41.9,\,-37.3]}
\)} \\
& WF
& \(54.9_{\scriptscriptstyle[54.1,\,55.7]}\)
& \(57.8_{\scriptscriptstyle[56.0,\,59.6]}\)
& \\

\addlinespace[2pt]

\multirow{2}{=}{Fully usable}
& MB
& \(97.0_{\scriptscriptstyle[95.8,\,98.1]}\)
& \(96.7_{\scriptscriptstyle[95.4,\,98.1]}\)
& \multirow{2}{*}{\(
-42.8_{\scriptscriptstyle[-45.2,\,-40.5]}
\)} \\
& WF
& \(50.4_{\scriptscriptstyle[49.5,\,51.2]}\)
& \(54.0_{\scriptscriptstyle[52.2,\,55.8]}\)
& \\

\bottomrule
\end{tabularx}
\end{table}
The audits show higher agreement on MemoryBench than on WildFB, in
line with the feedback-usability analysis. Across both sources, the
results support the intended atomic decomposition, task-relative role
assignment, and response-conditioned Specialist targets.

\subsection{Full Task Composition and Feedback Usability Results}
\label{app:task-feedback-usability}

Let $H$ denote the task context up to and including the original user
request, before the logged assistant response.

\begin{table}[!t]
\centering
\caption{Task-operation distribution inferred from $H$ alone. Values
are percentages of analyzed root training tasks.}
\label{tab:task-composition-full}
\footnotesize
\setlength{\tabcolsep}{4.0pt}
\renewcommand{\arraystretch}{1.06}
\begin{tabular}{lrr}
\toprule
\textbf{Task operation} & \textbf{MemoryBench} & \textbf{WildFB} \\
\midrule
Generation and creation       & 53.4 & 25.5 \\
Editing and transformation    &  5.8 & 13.6 \\
Explanation and summarization &  6.0 &  6.9 \\
Analysis and reasoning        & 32.7 &  7.5 \\
Information seeking and QA    &  1.0 & 17.4 \\
Code and implementation       &  0.0 & 22.6 \\
Advice and planning           &  1.1 &  5.2 \\
Other/mixed/uncertain         &  0.1 &  1.4 \\
\bottomrule
\end{tabular}
\end{table}

The analysis manifest contains 845 MemoryBench and 14,094 WildFB root
tasks. Complete outputs are available for 834 and 13,866 tasks,
respectively; incomplete provider output is the only exclusion. The
standardized comparison uses seven shared task-operation by
verifiability strata, covering 790 MemoryBench and 5,599 WildFB root
tasks. The standardized results therefore describe the shared-support
task distribution rather than the full WildFB training distribution.
These details yield the values reported in
Figure~\ref{fig:task-feedback-usability}.

\subsection{Stage-wise Contributions}
\label{app:stage-wise-transfer}

Subtracting adjacent endpoints reproduces all
$\Delta_{\mathrm G}$ and $\Delta_{\mathrm S}$ values in
Table~\ref{tab:stage-transfer}. The corresponding decision counts also
recover the reported \texttt{APPLY} and edit rates. Edited never
exceeds \texttt{APPLY}; MMLU-Pro is the all-\texttt{KEEP} case and
shows no Specialist increment.

\subsection{Observed Specialist Behavior on a Benchmark Task}
\label{app:specialist-case-study}

Figure~\ref{fig:specialist-case} presents a case study on AlpacaEval~2.0. The benchmark instruction is reproduced verbatim, while the Generalist response, Specialist output, integrated response, and pairwise outcome are taken directly from the corresponding evaluation run.

\end{document}